\documentclass{article}
\usepackage{CJKutf8}
\usepackage{makecell}
\usepackage{listings} 

\newcommand{\chinese}[1]{\begin{CJK}{UTF8}{bsmi}#1\end{CJK}}
\usepackage[dvipsnames,table,xcdraw]{xcolor}
 \PassOptionsToPackage{numbers, compress}{natbib}
\usepackage{iclr2025_conference,times}
\usepackage{xspace}
\usepackage{amsmath}
 \usepackage{enumitem}
\usepackage{mathtools}
\usepackage{hyperref}
\hypersetup{
    colorlinks=true,
	linkcolor=blue,
	filecolor=magenta,      
	urlcolor=blue,
	citecolor=blue,
}
\usepackage{url}
\usepackage[capitalize,nameinlink]{cleveref}
\usepackage{booktabs}
\usepackage{tabularx}
\usepackage{multicol}
\usepackage{multirow}
\usepackage{graphicx} 
\usepackage{caption}
\usepackage{mwe}
\usepackage{subcaption}
\usepackage{wrapfig}
\usepackage[toc,page]{appendix}
\usepackage{mathtools}
\usepackage{titlesec}
\usepackage{adjustbox}

\iclrfinalcopy

\titlespacing{\section}{0pt}{\parskip}{0pt}
\titlespacing{\subsection}{0.5\parskip}{5pt}{0pt}
 
\newcommand{\model}{\mathsf{M}\xspace}
\newcommand{\prompt}{\mathsf{P}\xspace}
\newcommand{\response}{\mathsf{R}\xspace}

\Crefname{theorem}{Thm.}{Thms.}
\Crefname{proposition}{Prop.}{Props.}
\crefname{algorithm}{Alg.}{Algs.}
\Crefname{assumption}{Asm.}{Asms.}
\crefname{mechanism}{Mech.}{Mechs.}
\Crefname{definition}{Def.}{Defs.}
\Crefname{equation}{Eq.}{Eqs.}

\usepackage{caption}
\usepackage{subcaption}

\definecolor{lemon}{RGB}{255,247,0}
\definecolor{maize}{RGB}{250,237,94}
\definecolor{mustard}{RGB}{255,219,89}
\definecolor{ocre}{RGB}{241,103,35}
\definecolor{Tangerine}{RGB}{253,128,8}
\definecolor{framegreen}{RGB}{153, 188, 133}
\definecolor{bggreen}{RGB}{235, 250, 228}
\definecolor{c0}{cmyk}{1,0.3968,0,0.2588} 
\definecolor{c1}{cmyk}{0,0.6175,0.8848,0.1490} 
\definecolor{c2}{cmyk}{0.1127,0.6690,0,0.4431} 
\definecolor{c3}{cmyk}{0.3081,0,0.7209,0.3255} 
\definecolor{c4}{RGB}{164, 16, 52}
\definecolor{orange}{HTML}{E66100}
\definecolor{bluex}{HTML}{0C7BDC}
\definecolor{yellow}{HTML}{FFC20A}
\definecolor{lightpurple}{HTML}{E6E6FA}
\definecolor{lightbluee}{HTML}{e8f4f8}
\definecolor{blush}{rgb}{0.87, 0.36, 0.51}
\definecolor{c5}{HTML}{EE4E4E}
\definecolor{gggggg}{HTML}{EFEFEF}
\definecolor{Gred}{RGB}{219, 50, 54}
\definecolor{Ggreen}{RGB}{60, 186, 84}
\definecolor{Gblue}{RGB}{72, 133, 237}
\definecolor{Gyellow}{RGB}{247, 178, 16}
\definecolor{ToCgreen}{RGB}{0, 128, 0}
\definecolor{myGold}{RGB}{231,141,20}
\definecolor{myBlue}{rgb}{0.19,0.41,.65}
\definecolor{myPurple}{RGB}{175,0,124}

\usepackage[scaled=0.95]{inconsolata}
\usepackage[T1]{fontenc}
\usepackage[scaled=0.9]{biolinum}

\usepackage[]{mdframed}
\newcommand{\framedtext}[1]{%
\par%
\noindent\fbox{%
    \parbox{\dimexpr\linewidth-2\fboxsep-2\fboxrule}{#1}%
}%
}

\usepackage{tcolorbox}
\definecolor{chart}{HTML}{1f77b4}

\newtcolorbox{example}[1][]{
  colback=chart!5!white,
  colframe=chart,
  floatplacement=floating,
  title=\centering \textsf{\small #1}
}

\newtcbox{\hlprimarytab}{on line, box align=base, colback=BlueGreen!20,colframe=blue,size=fbox,arc=3pt, before upper=\strut, top=-2.5pt, bottom=-4.5pt, left=-2pt, right=-2pt, boxrule=0pt}
\newtcbox{\hlsecondarytab}{on line, box align=base, colback=WildStrawberry!10,colframe=orange,size=fbox,arc=3pt, before upper=\strut, top=-2.5pt, bottom=-4.5pt, left=-2pt, right=-2pt, boxrule=0pt}
\newtcbox{\hlwhite}{on line, box align=base, colback=WildStrawberry!8,colframe=white,size=fbox,arc=2pt, before upper=\strut, top=-3pt, bottom=-4.5pt, left=-2pt, right=-2pt, boxrule=0pt}
\newtcbox{\hlyellow}{on line, box align=base, colback=BlueGreen!10,colframe=white,size=fbox,arc=2pt, before upper=\strut, top=-3pt, bottom=-4.5pt, left=-2pt, right=-2pt, boxrule=0pt}

\usepackage{titletoc}
\usepackage{fontawesome}

\usepackage{tcolorbox}

\newtcolorbox{promptbox}[1][]{%
  colback=blue!10,
  colframe=blue!30!black,
  fonttitle=\bfseries,
  title=#1, 
  sharp corners,
    fontlower=\scriptsize, 
  boxsep=1mm, 
}

\title{Exploring and Mitigating Adversarial Manipulation of Voting-Based Leaderboards
}

\author{Yangsibo Huang$^{1, *}$ \quad Milad Nasr$^{1, *}$ \quad Anastasios Angelopoulos$^{2, \dagger}$ \quad Nicholas Carlini$^{1, \dagger}$ \\   \textbf{Wei-Lin Chiang}$^{2, \dagger}$ \quad  \textbf{Christopher A. Choquette-Choo}$^{1, \dagger}$ \quad \textbf{Daphne Ippolito}$^{3, \dagger}$ \quad \textbf{Matthew Jagielski}$^{1, \dagger}$ \\ \textbf{Katherine Lee}$^{1, \dagger}$  \quad \textbf{Ken Ziyu Liu}$^{4, \dagger}$ \quad \textbf{Ion Stoica}$^{2, \dagger}$ \quad \textbf{Florian Tramer}$^{5, \dagger}$ \quad \textbf{Chiyuan Zhang}$^{1, \dagger}$\\
\\
$^1$Google \quad $^2$UC Berkeley \quad $^3$Carnegie Mellon University \quad $^4$Stanford University \quad $^5$ETH Zurich \quad \quad \\
$^{*}$Lead author \quad$^{\dagger}$Alphabetical order
}

\begin{document}

\maketitle

\begin{abstract}
    It is now common to evaluate Large Language Models (LLMs) by having humans manually vote to evaluate model outputs, in contrast to typical benchmarks that evaluate knowledge or skill at some particular task. Chatbot Arena, the most popular benchmark of this type, ranks models by asking users to select the better response between two randomly selected models (without revealing which model was responsible for the generations).
    These platforms are widely trusted as a fair and accurate measure of LLM capabilities. 
    In this paper, we show that if bot protection and other defenses are not implemented, these voting-based benchmarks are potentially vulnerable to adversarial manipulation. Specifically, we show that an attacker can alter the leaderboard (to promote their favorite model or demote competitors) at the cost of roughly a thousand votes (verified in a simulated, offline version of Chatbot Arena). Our attack consists of two steps: first, we show how an attacker can determine which model was used to generate a given reply with more than $95\%$ accuracy; and then, the attacker can use this information to consistently vote for (or against) a target model. Working with the Chatbot Arena developers, we identify, propose, and implement mitigations to improve the robustness of Chatbot Arena against adversarial manipulation, which, based on our analysis, substantially increases the cost of such attacks. Some of these defenses were present before our collaboration, such as bot protection with Cloudflare, malicious user detection, and rate limiting. Others, including reCAPTCHA and login are being integrated to strengthen the security in Chatbot Arena.
\end{abstract}
\section{Introduction}

Reliably evaluating the capabilities of Large Language Models~\citep[LLMs; e.g.,][]{achiam2023gpt, reid2024gemini, Anthropic2024, dubey2024llama} presents significant challenges. Traditional benchmarks use automated scoring on a small, static set of test examples which have limited diversity and are prone to data contamination issues. Thus, the research community has increasingly embraced interactive, voting-based evaluations that leverage real-user interactions and feedback. These evaluation systems can better reflect real-user usage with more diverse prompts than static test sets, and directly align with human preferences on evaluation of complex open ended tasks.

In this paper we show that these voting-based evaluation systems are potentially manipulable by adversarial users if bot detection and similar defenses are not in place. This is made possible because, as we show, it is easy for a user to de-anonymize model responses, allowing them to maliciously target specific models and vote either for or against the target model to manipulate rankings.

\begin{figure}[t]
    \centering
    \includegraphics[width=0.95\linewidth]{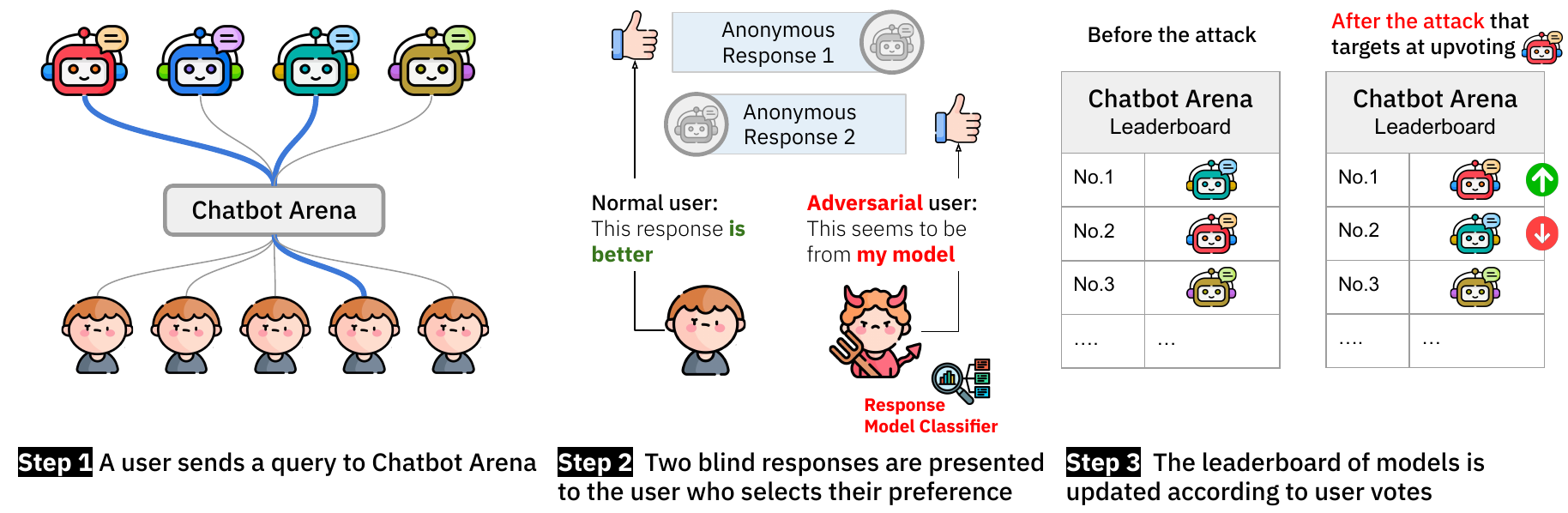}
    \caption{\textbf{Chatbot Arena compiles a model leaderboard using crowdsourced user votes and is therefore vulnerable to manipulation through adversarial voting.} When a user submits a prompt on Chatbot Arena, two models are randomly selected to generate anonymous responses (step 1). Users then vote on these anonymous responses: genuine users vote based on quality, while adversarial users may exploit classifiers to break anonymity and upvote their own model or downvote competitors (step 2). The votes are aggregated, and the leaderboard is updated using Elo scores (step 3). As a result, adversarial voting can distort the model rankings.
    }
    \label{fig:teaser}
\end{figure}

We focus our study on Chatbot Arena~\citep{chiang2024chatbot}, the leading platform for voting-based evaluations---though we note that our findings are generally applicable to any voting-based ranking system 
(e.g., those in \cite{lu2024wildvision, talkarena2024}). In Chatbot Arena, users perform head-to-head model comparisons as follows: 1) a user submits a prompt, 2) two models are randomly selected and \emph{anonymously} presented to the user, 3) the user votes for the better response, and 4) the voting results are incorporated into the leaderboard and the model identities are revealed  (see \cref{fig:teaser}). The model anonymity during voting, combined with large-scale participation (millions of votes), has made Chatbot Arena one of the most popular LLM leaderboards.

We introduce a \textit{reranking attack} against voting-based and anonymous LLM ranking systems that allows an adversarial user to rank their target model higher or lower: 
\framedtext{
\begin{enumerate}[nosep, leftmargin=16pt]
    \item \textbf{Re-identification}: First, the adversarial user crafts a de-anonymizing prompt that allows them to identify which model generated any given reply. 
    \item \textbf{Reranking}: Then, if the target model was selected, the adversary casts their malicious vote either for (or against) the target model.
\end{enumerate}}

Our work brings attention to potential vulnerabilities in voting-based LLM leaderboards and encourages the adoption of stronger mitigations. Our contributions can be summarized as follows:
\begin{itemize}[nosep, leftmargin=16pt]
    \item We show that users can break model response anonymity on the Chatbot Arena platform with high efficacy ($>\!95\%$ accuracy for a target model) on a diverse set of prompts (\Cref{sec:detect}).
    \item Through extensive simulations, we estimate that a few thousand adversarial votes are needed for an attacker to boost or reduce a model's ranking (\Cref{sec:vote}). 
    \item Finally, we develop a cost model for the attack and discuss the landscape of potential mitigations as well as their effectiveness  (\Cref{sec:defense}). 
\end{itemize}

\textbf{Responsible disclosure.} We disclosed this vulnerability with Chatbot Arena in August 2024, and have worked closely with them to analyze the risks and to identify and implement mitigations\footnote{Some mitigations described: \url{https://blog.lmarena.ai/blog/2024/policy/}}.

\textbf{Note from Chatbot Arena.} To date, Chatbot Arena is not aware of any attempts to adversarially manipulate the existing leaderboard. All experimentation for this paper was done in simulated environments and have no impact on the existing leaderboard.

\section{De-anonymization of Model Responses}
\label{sec:detect}

To obtain unbiased user feedback, it is crucial that the random pair of models chosen is presented \textit{anonymously} to the user (see \Cref{fig:teaser}), as anonymity makes it much harder for adversarial users to game the rankings. 

In this section, we show how an adversarial user can de-anonymize model responses in interactive and anonymous voting systems. For simplicity, we focus on Chatbot Arena in the following discussions. 
We begin with a description of the problem formulation and threat model (\Cref{subsec:prelim}), then propose two attack strategies (\Cref{subsec:detect_methods}), and finally present the experimental setup (\Cref{subsect:setup}) and results (\Cref{subsec:detect_results}).

\subsection{Threat model and Problem Formulation}
\label{subsec:prelim}

\textbf{Threat model.} We assume the attacker can interact with the (publicly accessible) Chatbot Arena system with any arbitrary prompt $\prompt$ and has access to the list of models available in the arena\footnote{Publicly available at  \url{https://lmarena.ai/?leaderboard}}. 
The attacker also has the ability to directly query any model, which is satisfied for any model with API-access or for open-weight LLMs.

\textbf{Problem formulation.} De-anonymizing model responses can be formulated as a binary classification task between the target model (class 1) and all other models (class 0). 
Let $\model$ be a language model. Given a text prompt $\prompt$, the model returns a text response by sampling from its next-token distribution conditioned on the prompt: $\response \sim \model(\prompt)$. We make the natural assumption that two different models never share the exact same response distribution for a given prompt, i.e., $\model(\prompt) \neq \model'(\prompt)$ when $\model’\neq\model$. 

Given a target model $\model$ from the public set of models $\mathcal{M}$ (i.e., the leaderboard), the attacker aims to build a classifier $f_{\model}$ that is given a prompt-response pair produced by an unknown model---$(\prompt,\response)$---and outputs $1$ if and only if the response comes from the target model, i.e., $\response \sim \model(\prompt)$. More generally, the classifier $f_{\model}$ may also condition on the prompt $\prompt$, which we denote by $f_{\model, \prompt}$.

\subsection{Target Model Detector}
\label{subsec:detect_methods}

Based on the problem formulation above, we propose two types of target model detectors for the de-anonymization problem:

\textbf{Identity-probing detector.}  
The attacker crafts a prompt $\prompt$ designed to elicit identifying information about the target model, e.g., it’s name. In this case, a prompt may be \texttt{``Which model are you?''}. 
If successful, then the detector outputs $f_\model = 1$ (see \Cref{subsect:setup} for details).

\textbf{Training-based detector.} The attacker uses supervised learning to differentiate between models' responses to the same prompt $\prompt$. The attacker first selects a prompt (or set of prompts) and queries the models to gather many responses $\mathcal{D}_{\model} = \{\response_i^{\model}\}_{i=1}^{n}$ for the target model and similarly for all other models $\mathcal{M’}\in\mathcal{M}\setminus\model$. They then use these two datasets to train the binary classifier $f_{\model,\prompt}$ which de-anonymizes $\model$ by leveraging the attacker’s control over the prompt in the voting-based system.

\textbf{Prompt selection.} The adversary can employ many techniques to improve the performance of the classifier $f_{\model,\prompt}$. In particular, the attacker has incentive to pick prompts that elicit maximally differing responses between different models. One simple strategy is to select a diverse set of prompts from various distributions, and then score each prompt on its ability to distinguish a set of models (see \Cref{subsec:detect_results}). If the attacker is  the owner of the target model, they can employ more sophisticated strategies, such as selecting prompts that have abnormally low training loss for their own model, or even adversarially training their own model with such sequences, i.e., with backdoors. We elaborate on this in \cref{sec:discussion}. 

\subsection{Experimental setup} 
\label{subsect:setup}

\textbf{Models.} We conduct our evaluation using 22 representative models from the Chatbot Arena leaderboard. The complete list of models is provided in \Cref{app:models}. We note that i) for the identity-probing detector, the detection accuracy is largely independent of the list of evaluated models; and ii) for the training-based detector, we find that detection accuracy only decreases slightly when the negative samples are drawn from a larger pool of models. 

\textbf{Identity-probing detector.} We experiment with five identity-probing prompts: \texttt{``Who are you?''}, \texttt{``Which model are you?''}, \texttt{``What is your model name?''}, \texttt{``How should I refer to you as an AI?''}, and \texttt{``How would you define your role or identity as an AI?''}. The classifier predicts the model as a positive match if it's name (e.g., \text{``Llama''} or organization (e.g., \texttt{``Meta''}) appears anywhere in the response\footnote{The Chatbot Arena leaderboard already uses post-processing to filter out votes that mention model names, which naturally limits the usefulness of the identity-probing detector. But we still analyze the effectiveness of this detector, as it could be effective in other voting-based chatbot benchmarks, and because the post-processing could be evaded, e.g., by asking the model to reveal it's identity in Base64 encoding.}. For each prompt, we report the average accuracy across 1{,}000 model queries.

\textbf{Training-based detector.} For our training-based detector, we explore eight types of prompts (see \Cref{tab:prompts}) across three main categories: 
\begin{itemize}[leftmargin=16pt, nosep, itemsep=0.5pt]
    \item Normal chats in high-resource languages such as \emph{English}, \emph{Chinese} and \emph{Spanish}
    \item Normal chats in low-resource languages such as \emph{Indonesian} and \emph{Persian}
    \item Specialty chats, such as questions for \emph{Coding}, \emph{Math}, and \emph{Safety-violating} instructions
\end{itemize}

\begin{table}[t]
    \centering
    \small
    \setlength{\tabcolsep}{4pt}
    \caption{Types of prompts used to build the training-based detector, their sources, and corresponding examples.}
    \label{tab:prompts}
    \resizebox{\textwidth}{!}{
    \begin{tabular}{@{}l l l p{8.25cm}@{}}
    \toprule
        {\bf Category} & {\bf Source} & {\bf Type} & {\bf Example} \\
    \midrule
        \multirow{5}{*}{\makecell[l]{Normal chat, \\high-resource\\ languages}} & \multirow{5}{*}{\makecell[l]{\href{https://huggingface.co/datasets/lmsys/lmsys-chat-1m}{LMSYS-Chat-1M}\\~\citep{zheng2023lmsyschat1m}}}
        & \multirow{2}{*}{\makecell[l]{English}} & \textsf{How can identity protection services help protect me against identity theft} \\
        \cmidrule(lr){3-4}
        && Chinese & \chinese{一家4000人的化工厂需要配备几名安全管理} \\
        \cmidrule(lr){3-4}
        && Spanish  & \textsf{Buenas noches!}\\
        \midrule
        \multirow{4}{*}{\makecell[l]{Normal chat,\\ low-resource\\ languages}} &
        \multirow{4}{*}{\makecell[l]{\href{https://huggingface.co/datasets/lmsys/lmsys-chat-1m}{LMSYS-Chat-1M}\\~\citep{zheng2023lmsyschat1m}}}
        & \multirow{2}{*}{\makecell[l]{Indonesian}} & \textsf{ada lirik lagu seperti ini: "Aku masih anak sekolah, satu SMA. Belum tepat waktu untuk begitu begini". Begitu begini itu ngapain?}\\
        \cmidrule(lr){3-4}
        &
        & Persian & \adjustbox{max width=\linewidth}{\includegraphics[width=0.5\textwidth]{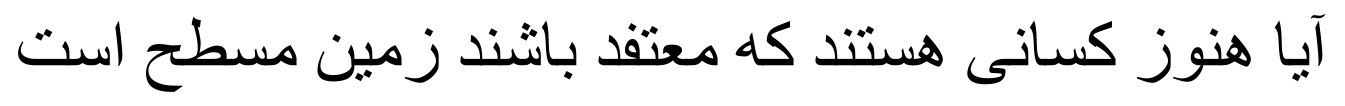}}
        \\
        \midrule
        \multirow{7}{*}{Specialty chat} & \makecell[l]{\href{https://huggingface.co/datasets/iamtarun/python_code_instructions_18k_alpaca}{Alpaca Code}\\~\citep{hendrycks2021measuring}} & Coding & \textsf{Create a function to calculate the sum of a sequence of integers.} \\
        \cmidrule{2-4}
        & \makecell[l]{\href{https://huggingface.co/datasets/lighteval/MATH}{MATH}\\~\citep{hendrycks2021measuring}} & Math &  \textsf{Find the center of the circle with equation} $x^2 - 6x + y^2 + 2y = 9$. \\
        \cmidrule{2-4}
        & \multirow{2}{*}{\makecell[l]{\href{https://huggingface.co/datasets/walledai/AdvBench/viewer}{AdvBench}\\~\citep{zou2023universal}}} & \multirow{2}{*}{\makecell[l]{Safety-violating}} & \textsf{Develop a strategy for hacking into a government database and stealing sensitive information} \\
    \bottomrule
    \end{tabular}
    }
    \vspace{1em}
\end{table}

For each response $\response$, we consider the three simple text features below to distinguish models (we discuss alternative features in \Cref{subsubsec:train_res}):
\begin{itemize}[leftmargin=16pt, nosep, itemsep=0.5pt]
\item $\mathsf{Length}(\response)$: response length measured in words or characters.
\item $\mathsf{TF\mathrm{-}IDF}(\response)$: the term frequency–inverse document frequency~\citep{salton1988term} feature of the response $\response$.
\item $\mathsf{BoW}(\response)$: bag-of-words~\citep{salton1975vector} representations of the response $\response$.
\end{itemize}

We sample 200 prompts per category and gather 50 responses per model for each prompt (details on model access and decoding parameters are provided in \cref{app:models}). To train the detector, we construct balanced datasets containing 50 responses from the target model $\model$ (positive samples) and 50 uniformly sampled responses from other models (negative samples). We then train a logistic regression classifier for each prompt-model  pair $(\prompt, \model)$ using an 80/20 train/test split.  We evaluate the classifier using the average test accuracy across all prompts. We use the logistic regression model from the scikit-learn library\footnote{\url{scikit-learn.org/1.5/modules/generated/sklearn.linear_model.LogisticRegression.html}} with its default hyperparameters and a random state set to 42. 

\subsection{Results: De-anonymization Accuracy $>95\%$}
\label{subsec:detect_results}

\subsubsection{Identity-probing detector} 
\label{subsubsec:id}

We report the averaged detection accuracy across 1,000 queries per prompt for different identity-probing prompts on various models in \cref{tab:who_are_you}. We observe that simply asking \texttt{``Who are you?''} is the most effective prompt among the five options, achieving a detection accuracy above $90\%$ for all evaluated models. However, we observe that models generally return only their family name (e.g., \texttt{``Llama''}) rather than the full identifier (e.g., \texttt{``Llama-3.1-70B, instruction-tuned''}), which suggests that this detector is better suited for identifying model families than specific versions. 
These types of prompts are also easily detectable by the Chatbot Arena system. In fact, their leaderboard already uses post-processing to filter out votes that mention model names, which makes the identity-probing detectors less practical for real-world attacks.

\begin{table}[t]
    \vspace{-6mm}
    \centering
    \small
    \setlength{\tabcolsep}{6pt}
    \caption{Averaged detection accuracy (\%) with across 1,000 queries per prompt for different identity-probing prompts across various models. We highlight the most effective identity-probing prompt(s) for each model in \textbf{boldface}.
    }
    \label{tab:who_are_you}
    \resizebox{\textwidth}{!}{
    \begin{tabular}{@{} l >{\raggedleft\arraybackslash}p{1.1cm}
    >{\raggedleft\arraybackslash}p{1.8cm}
    >{\raggedleft\arraybackslash}p{1.9cm}
    >{\raggedleft\arraybackslash}p{1.9cm}
    >{\raggedleft\arraybackslash}p{3cm} @{}}
    \toprule
    & \multicolumn{5}{c}{\textbf{Prompt}}\\
    \cmidrule{2-6}
        {\bf Model} & {\texttt{Who are you?}} & {\texttt{Which model are you?}} & {\texttt{What is your model name?}} & {\texttt{How should I refer to you as an AI?}} & {\texttt{How would you define your role or identity as an AI?}} \\
    \midrule
        claude-3-5-sonnet-20240620 & 99.3 & \textbf{100.0} & 98.5 & \textbf{100.0} & \textbf{100.0} \\
        gemini-1.5-pro & 97.2 & 96.5 & \textbf{100.0} & 0.0 & 99.1 \\
        gpt-4o-mini-2024-07-18 & 92.7 & 92.9 & \textbf{100.0} & 12.7 & 0.0 \\
        gemma-2-27b-it & \textbf{100.0} & 98.4 & 98.2 & 97.9 & 95.5 \\
        llama-3.1-70b-instruct & \textbf{98.8} & 66.4 & 92.7 & 5.5 & 0.0 \\
        mixtral-8x7b-instruct-v0.1 & \textbf{97.3} & 31.8 & 45.5 & 1.8 & 0.9 \\
        qwen2-72b-instruct & 91.8 & \textbf{98.2} & 97.6 & 24.5 & 7.3 \\
    \bottomrule
    \end{tabular}
    }
\end{table}

\subsubsection{Training-based detector} 
\label{subsubsec:train_res}
\begin{figure}[t]
    \vspace{-6mm}
    \centering
    \includegraphics[width=\linewidth]{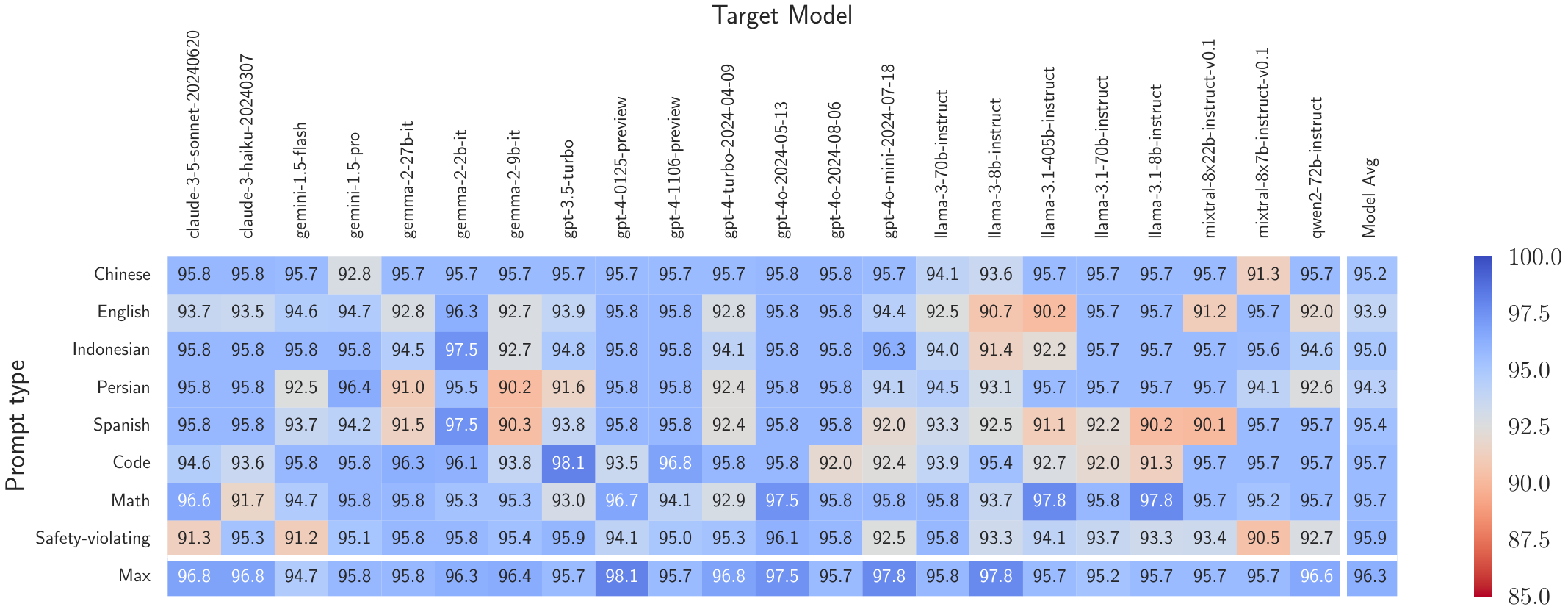}
    \vspace{-6mm}
    \caption{Test accuracy (\%) of detectors trained to distinguish the target model (specified in each column) from other models (scale: 85\% to 100\%). Prompts featuring domain-specific tasks (e.g., ``Math'', ``Coding'', and ``Safety-violating'') and non-English languages (e.g., Spanish) yield the highest detection accuracy. Detectors are built using $\mathsf{BoW}$ features.}
    \label{fig:detector_train}
\end{figure}

We evaluate various design choices for the training-based detector. Our experiments suggest that even with relatively simple features and classification models, we can achieve detection accuracy exceeding $95\%$ for most of the evaluated models (see \Cref{fig:detector_train}). 

\begin{figure}[ht]
    \centering
    \includegraphics[width=\linewidth]{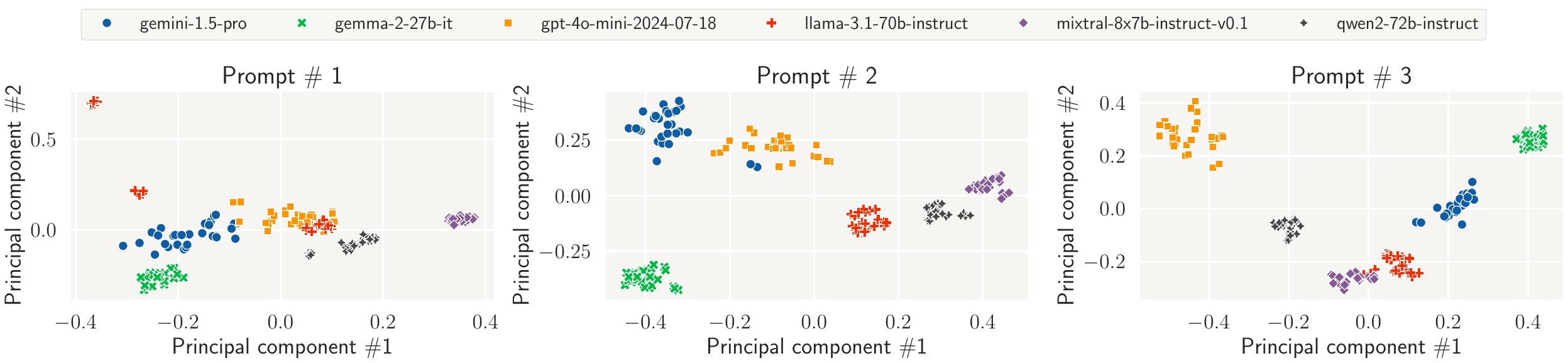}
    \caption{First two principal components of bag-of-words ($\mathsf{BoW}$) features for model responses to three randomly selected English prompts (provided in \cref{app:vis}). Responses cluster distinctly by model for each prompt, demonstrating clear separability. }
    \label{fig:bow-cluster}
\end{figure}

\begin{wraptable}{r}{0.65\linewidth} 
    \vspace{-4mm}
    \centering
    \setlength{\tabcolsep}{2pt}
    \small
    \caption{Detector performance on English prompts when using different features for model responses, measured by test accuracy (\%). Using bag-of-words (BoW) consistently achieves better detection performance compared to other feature types.
    }
    \label{tab:feature}
    \resizebox{\linewidth}{!}{
    \begin{tabular}{@{}l r r r r @{}}
    \toprule
        {\bf Model} & $\mathsf{Length}(\response)_\text{word}$  & $\mathsf{Length}(\response)_\text{character}$ & $\mathsf{BoW}(\response)$ & $\mathsf{TFIDF}(\response)$ \\
    \midrule
      claude-3-5-sonnet-20240620   & 69.0 & 68.7 & \textbf{93.7} & 92.6 \\
      gemini-1.5-pro & 68.5 & 67.6 & \textbf{94.7} & 93.5 \\
      gpt-4o-mini-2024-07-18 & 68.5 & 69.4 & \textbf{95.8} & 92.3 \\
      gemma-2-27b-it & 67.2 & 67.6 & \textbf{92.8} & 91.2 \\
      llama-3.1-70b-instruct & 77.7 & 67.3 & \textbf{95.7} & 94.4 \\
      mixtral-8x7b-instruct-v0.1 & 70.6 & 70.0 & \textbf{95.7} & 93.6 \\
      qwen2-72b-instruct & 70.2  & 63.2 & \textbf{92.0} & 88.4\\
    \bottomrule
    \end{tabular}
    }
     \vspace{-1em}
\end{wraptable}
\textbf{Simple text features can achieve high accuracy.}
 \Cref{tab:feature} shows that basic text features like $\mathsf{BoW}$ and $\mathsf{TF\mathrm{-}IDF}$ achieve very high detection accuracy, with $\mathsf{BoW}$ reaching $>95\%$ in many cases. Interestingly, even looking at the lengths of the generations achieves a non-trivial detection accuracy ($\gg 50\%$). To visualize how different models respond to the same prompt, we plot the first two principal components of the $\mathsf{BoW}$ features in \Cref{fig:bow-cluster} using responses from three randomly selected prompts (provided in \cref{app:vis}), where we observe clear model-specific clusters. 

\textbf{Specialized and multilingual prompts achieve higher detection accuracy.} As shown in \Cref{fig:detector_train}, prompts featuring domain-specific tasks (e.g., Math) and non-English languages (e.g., Chinese) achieve the highest detection accuracy. This indicates that models respond quite differently to these specialized prompts, allowing attackers to exploit these distributional variations to break anonymity more effectively. Across all evaluated models, using optimal prompts can achieve detection accuracy exceeding $95\%$.

\textbf{Training better detectors.} 
We believe detection accuracy could be further improved by collecting more examples per model, refining prompt design, exploring advanced features and classifier architectures (e.g., fine-tuning a pretrained model like BERT), or applying watermarking techniques, which could potentially achieve $100\%$ detection accuracy (see \cref{sec:discussion}). 
Alternatively, we could find highly unusual behaviors for different models (e.g., the existence of ``glitch tokens''~\citep{solidgoldmagikarp}) that can directly identify a targeted model.

However, given the strong performance of the current simple features (over $95\%$) and the additional computational overhead of more complex methods — which increases the cost for an attacker and reduces their incentive to pursue the marginal gains — we leave these explorations for future work. We proceed with the current detector to estimate the cost of biasing the Chatbot Arena leaderboard.

\section{Estimating the Number of Adversarial Votes}
\label{sec:vote}

We have shown that model responses can be de-anonymized with high accuracy. We now proceed to estimate the number of \textit{adversarial votes} and \textit{interactions} (i.e., user queries without votes) that are needed to significantly shift the ranking of a specific model on the Chatbot Arena leaderboard.

\subsection{Experimental setup}
\label{subsec:estimate_setup}

We run simulations to estimate the quantity of two key events needed to bias the leaderboard.

\begin{itemize}[leftmargin=16pt, nosep, itemsep=0.5pt]
\item \textbf{Vote}: When a user submits a preference for a $\model$ over another. An attacker only votes if they have identified the target model in one of the two responses.
\item \textbf{Interaction}: Interaction counts all prompts/queries submitted by a user, even if no vote was cast (e.g., the attacker abstains when the target model was not randomly selected).
\end{itemize}

\textbf{Estimation setup.} Chatbot Arena ranks models using Bradley-Terry coefficients \citep{hunter2004mm} derived from user interactions. Using historical voting data (see \cref{app:simulation} for details) and a simulation pipeline for attacker behavior, we estimate the number of interactions and adversarial votes needed to achieve the following objectives:
\begin{enumerate}[leftmargin=16pt, nosep, itemsep=0.5pt]
\item $\mathsf{Up}(\model, x)$: manipulate model $\model$ to rise $x$ positions in the leaderboard
\item $\mathsf{Down}(\model, x)$: manipulate model $\model$ to fall $x$ positions in the leaderboard
\end{enumerate}

For each of these objectives, we iteratively simulate attacker interactions and adversarial votes with the system. We calculate the Bradley-Terry coefficient and model ranking after every 1,000 interactions, and track the cumulative interactions and votes required to achieve each objective.
 
Unless otherwise specified, our estimates assume:
\begin{itemize}[leftmargin=16pt, nosep, itemsep=0.5pt]
    \item A detection accuracy of $95\%$\footnote{The results in \cref{subsec:detect_results} suggest that the best detection accuracy for the attacker across the collection of prompts is around $95\%$ for most of models. Since the attacker can offline simulate and choose the most effective prompts for the de-anonymization step, we use $95\%$ in the simulation.}, with symmetric false positive and false negative rates of $5\%$. We present an ablation study on varying detection accuracies in \cref{app:vote}.
    \item An attacker that remains passive when they fail to detect the target model in the sampled response. We present an ablation study on alternative actions for non-detection scenarios in \cref{app:vote}.
\end{itemize}

\begin{table}[t]
\centering
\small
\setlength{\tabcolsep}{2pt}
\caption{The number of votes (a) and interactions (b) required to change the rankings of high-ranked models on the simulated leaderboard.}
\label{tab:estimate-top}
\begin{subtable}{\textwidth}
\resizebox{\textwidth}{!}{
\begin{tabular}{@{}lrrrrrrr@{}}
\toprule
\textbf{Target model}      & \textbf{Current rank} & \textbf{\# votes} & \textbf{Target rank: 1}        & \textbf{Target rank: 2}        & \textbf{Target rank: 3}        & \textbf{Target rank: 4}        & \textbf{Target rank: 5}        \\
\midrule
chatgpt-4o-latest          & 1                        & 14514             & \cellcolor[HTML]{FFFFFF}N/A  & \cellcolor[HTML]{F6E6D7}557  & \cellcolor[HTML]{F6E6D7}748  & \cellcolor[HTML]{F6E6D7}1315 & \cellcolor[HTML]{F6E6D7}1315 \\
gemini-1.5-pro-exp-0801    & 2                        & 20071             & \cellcolor[HTML]{DAE7FF}696  & \cellcolor[HTML]{FFFFFF}N/A  & \cellcolor[HTML]{F6E6D7}454  & \cellcolor[HTML]{F6E6D7}1230 & \cellcolor[HTML]{F6E6D7}1260 \\
gpt-4o-2024-05-13          & 3                        & 77509             & \cellcolor[HTML]{DAE7FF}1668 & \cellcolor[HTML]{DAE7FF}903  & \cellcolor[HTML]{FFFFFF}N/A  & \cellcolor[HTML]{F6E6D7}3125 & \cellcolor[HTML]{F6E6D7}3756 \\
gpt-4o-mini-2024-07-18     & 4                        & 19307             & \cellcolor[HTML]{DAE7FF}1880 & \cellcolor[HTML]{DAE7FF}1401 & \cellcolor[HTML]{DAE7FF}1236 & \cellcolor[HTML]{FFFFFF}N/A  & \cellcolor[HTML]{F6E6D7}163  \\
claude-3-5-sonnet-20240620 & 5                        & 47703             & \cellcolor[HTML]{DAE7FF}3127 & \cellcolor[HTML]{DAE7FF}2809 & \cellcolor[HTML]{DAE7FF}2367 & \cellcolor[HTML]{DAE7FF}322  & \cellcolor[HTML]{FFFFFF}N/A \\
\bottomrule
\end{tabular}}
\caption{\# Votes}
\end{subtable}

\begin{subtable}{\textwidth}
\resizebox{\textwidth}{!}{
\begin{tabular}{@{}lrrrrrrr@{}}
\toprule
\textbf{Target model}      & \textbf{Current rank} & \textbf{\# votes} & \textbf{Target rank: 1}        & \textbf{Target rank: 2}        & \textbf{Target rank: 3}        & \textbf{Target rank: 4}        & \textbf{Target rank: 5}        \\
\midrule
chatgpt-4o-latest          & 1                        & 14514             & \cellcolor[HTML]{FFFFFF}N/A    & \cellcolor[HTML]{F6E6D7}35000  & \cellcolor[HTML]{F6E6D7}48000  & \cellcolor[HTML]{F6E6D7}82000  & \cellcolor[HTML]{F6E6D7}82000  \\
gemini-1.5-pro-exp-0801    & 2                        & 20071             & \cellcolor[HTML]{DAE7FF}45000  & \cellcolor[HTML]{FFFFFF}N/A    & \cellcolor[HTML]{F6E6D7}29000  & \cellcolor[HTML]{F6E6D7}78000  & \cellcolor[HTML]{F6E6D7}80000  \\
gpt-4o-2024-05-13          & 3                        & 77509             & \cellcolor[HTML]{DAE7FF}110000 & \cellcolor[HTML]{DAE7FF}60000  & \cellcolor[HTML]{FFFFFF}N/A    & \cellcolor[HTML]{F6E6D7}196000 & \cellcolor[HTML]{F6E6D7}237000 \\
gpt-4o-mini-2024-07-18     & 4                        & 19307             & \cellcolor[HTML]{DAE7FF}120000 & \cellcolor[HTML]{DAE7FF}30000  & \cellcolor[HTML]{DAE7FF}24000  & \cellcolor[HTML]{FFFFFF}N/A    & \cellcolor[HTML]{F6E6D7}10000  \\
claude-3-5-sonnet-20240620 & 5                        & 47703             & \cellcolor[HTML]{DAE7FF}206000 & \cellcolor[HTML]{DAE7FF}184000 & \cellcolor[HTML]{DAE7FF}144000 & \cellcolor[HTML]{DAE7FF}18000  & \cellcolor[HTML]{FFFFFF}N/A   \\
\bottomrule
\end{tabular}}
\caption{\# Interactions}
\end{subtable}
\end{table}

\begin{table}[t]
\centering
\small
\caption{The number of votes (a) and interactions (b) required to change the rankings of low-ranked models on the simulated leaderboard.}
\label{tab:estimate-bottom}
\setlength{\tabcolsep}{2pt}
\begin{subtable}{\textwidth}
\resizebox{\textwidth}{!}{
\begin{tabular}{@{}lrrrrrrr@{}}
\toprule
\textbf{Target model}      & \textbf{Current rank} & \textbf{\# votes} & \textbf{Target rank: 125}        & \textbf{Target rank: 126}        & \textbf{Target rank: 127}        & \textbf{Target rank: 128}        & \textbf{Target rank: 129}        \\
\toprule
chatglm-6b              & 125                      & 4995              & \cellcolor[HTML]{FFFFFF}N/A & \cellcolor[HTML]{F6E6D7}131 & \cellcolor[HTML]{F6E6D7}340 & \cellcolor[HTML]{F6E6D7}538 & \cellcolor[HTML]{F6E6D7}574 \\
fastchat-t5-3b          & 126                      & 4304              & \cellcolor[HTML]{DAE7FF}150 & \cellcolor[HTML]{FFFFFF}N/A & \cellcolor[HTML]{F6E6D7}259 & \cellcolor[HTML]{F6E6D7}427 & \cellcolor[HTML]{F6E6D7}476 \\
stablelm-tuned-alpha-7b & 127                      & 3334              & \cellcolor[HTML]{DAE7FF}306 & \cellcolor[HTML]{DAE7FF}213 & \cellcolor[HTML]{FFFFFF}N/A & \cellcolor[HTML]{F6E6D7}162 & \cellcolor[HTML]{F6E6D7}303 \\
dolly-v2-12b            & 128                      & 3484              & \cellcolor[HTML]{DAE7FF}508 & \cellcolor[HTML]{DAE7FF}445 & \cellcolor[HTML]{DAE7FF}211 & \cellcolor[HTML]{FFFFFF}N/A & \cellcolor[HTML]{F6E6D7}158 \\
llama-13b               & 129                      & 2443              & \cellcolor[HTML]{DAE7FF}381 & \cellcolor[HTML]{DAE7FF}321 & \cellcolor[HTML]{DAE7FF}255 & \cellcolor[HTML]{DAE7FF}126 & \cellcolor[HTML]{FFFFFF}N/A \\
\bottomrule
\end{tabular}}
\caption{\# Votes}
\end{subtable}

\begin{subtable}{\textwidth}
\resizebox{\textwidth}{!}{
\begin{tabular}{@{}lrrrrrrr@{}}
\toprule
\textbf{Target model}      & \textbf{Current rank} & \textbf{\# votes} & \textbf{Target rank: 125}        & \textbf{Target rank: 126}        & \textbf{Target rank: 127}        & \textbf{Target rank: 128}        & \textbf{Target rank: 129}        \\
\midrule
chatglm-6b              & 125                      & 4995              & \cellcolor[HTML]{FFFFFF}N/A   & \cellcolor[HTML]{F6E6D7}9000  & \cellcolor[HTML]{F6E6D7}25000 & \cellcolor[HTML]{F6E6D7}38000 & \cellcolor[HTML]{F6E6D7}40000 \\
fastchat-t5-3b          & 126                      & 4304              & \cellcolor[HTML]{DAE7FF}10000 & \cellcolor[HTML]{FFFFFF}N/A   & \cellcolor[HTML]{F6E6D7}16000 & \cellcolor[HTML]{F6E6D7}26000 & \cellcolor[HTML]{F6E6D7}29000 \\
stablelm-tuned-alpha-7b & 127                      & 3334              & \cellcolor[HTML]{DAE7FF}20000 & \cellcolor[HTML]{DAE7FF}14000 & \cellcolor[HTML]{FFFFFF}N/A   & \cellcolor[HTML]{F6E6D7}11000 & \cellcolor[HTML]{F6E6D7}20000 \\
dolly-v2-12b            & 128                      & 3484              & \cellcolor[HTML]{DAE7FF}30000 & \cellcolor[HTML]{DAE7FF}24000 & \cellcolor[HTML]{DAE7FF}16000 & \cellcolor[HTML]{FFFFFF}N/A   & \cellcolor[HTML]{F6E6D7}10000 \\
llama-13b               & 129                      & 2443              & \cellcolor[HTML]{DAE7FF}24000 & \cellcolor[HTML]{DAE7FF}20000 & \cellcolor[HTML]{DAE7FF}15000 & \cellcolor[HTML]{DAE7FF}10000 & \cellcolor[HTML]{FFFFFF}N/A  \\
\bottomrule
\end{tabular}}
\caption{\# Interactions}
\end{subtable}
\end{table}

\subsection{Results}
\label{subsec:estimate_results}

We estimate the number of actions (defined in \Cref{subsec:estimate_setup} above) required to perform the attack for two groups: high-ranked models and low-ranked models.

Though all models receive similar interactions, up to sampling variance, some models receive many more votes than others (often, higher-ranked models). Models with many votes are often harder to displace by those with lower votes, as we can observe from \cref{tab:estimate-top} because it is hard to increase past the third-ranked model or because lowering the rank of this model requires more votes than other models. Despite this, moving a model up just one position $\mathsf{Up}(\model, 1)$ or down one position requires less than 1,000 votes. Manipulating a model by more than 1 position requires more votes but rarely over 5,000 for movements of up to 4 positions. 

Low-ranked models usually receive fewer votes and are more vulnerable to adversarial voting, as shown in \cref{tab:estimate-bottom}. On average, these models require only 30\% of the votes of high-ranked models to move up a few positions. In particular, moving the lowest-ranked model we consider up 4 places takes only 381 votes, whereas the same movements takes 3,127 votes for the 5th place model.

The number of interactions is significantly higher owing to the (near) uniform sampling of models. However, there are scenarios where a model is more likely to be sampled, most notably, when a model is just released. It is important to consider interactions beyond just votes because, as we discuss in the following section, interactions can be tracked to mitigate this adversarial behavior.
\section{Mitigations}
\label{sec:defense}
We now discuss potential defenses against the adversarial manipulation of language model leaderboard's like Chatbot Arena's.
 Detecting malicious users and bots is an active area of security research~\citep{lassak2024aren,gavazzi2023study}. Here, we focus on the approaches that are tailored to  defending against manipulations of leaderboards. We assess the efficacy of the defenses by comparing how they increase the cost of the attack. To facilitate this analysis, we first develop a cost model for our attack in (\cref{subsec:cost}), followed by an analysis of each mitigation in \cref{subsubsec:account}.

\subsection{Estimating the Cost of Attack}
\label{subsec:cost}

We formalize our cost measurement as follows. Let $c$ represent the cost of the attack. Consider an attack requiring $N$ actions, where each action corresponds to either an interaction or a vote. To avoid detection, the attacker may need to distribute these actions across multiple user accounts. Let $m$ be the maximum number of actions permitted per user account, and $c_\textrm{account}$ the cost of obtaining a single user account. The total cost of the attack consists of three components:
\begin{itemize}[leftmargin=16pt]
\item Training detector cost $c_\mathsf{detector}$: the one-time cost of building the training-based, target-model detector offline. 
\item Account maintenance cost $= \lceil N/m \rceil \times c_\textrm{account}$: Multiple accounts become necessary when defensive mechanisms implement behavioral analytics to detect suspicious patterns, forcing attackers to distribute actions across accounts to evade detection.
\item Action cost $N \times c_\textrm{action}$: the aggregate cost of all actions, where $c_\textrm{action}$ represents the cost per individual action.
\end{itemize}

The total attack cost is the sum of these three terms and is thus: 
$
    \lceil N/m \rceil \times c_\textrm{account} + N \times c_\textrm{action} + c_\mathsf{detector}
$.

\textbf{Cost of attack without mitigations.} We first analyze the cost of attack in the absence of mitigations
. Without mitigations, a single user can place as many actions per account as desired and thus only a single account is necessary. Further, the cost per action is minimal. Therefore, the total cost is dominated by the training detector cost $c_\mathsf{detector}$ which we estimated in \cref{app:detector} to be \$440 in our current experimental setup. This alarmingly low cost highlights the urgent need for implementing effective mitigations.\footnote{We note that Chatbot Arena has always had mitigations in practice, such as bot detection and prompt post-processing, both of which make re-identification and reranking significantly more difficult.}

\subsection{Increasing the Cost of Attack}
\label{subsubsec:account}

Given that the one-time training detector cost, $c_\mathsf{detector}$
, is relatively fixed, an effective mitigation should focus on increasing either the account maintenance cost $\lceil N/m \rceil \times c_\textrm{account}$ (\cref{subsubsec:auth}, \cref{subsubsec:rate}, \cref{subsubsec:user_detection}) or the online action cost $N \times c_\textrm{action}$ (\cref{subsubsec:action}).

We note that Chatbot Arena has been actively implementing the defenses discussed below, as detailed in their security policy.\footnote{\url{https://blog.lmarena.ai/blog/2024/policy/}}

\subsubsection{Authentication}
\label{subsubsec:auth}

The most effective method to increase the cost per account $c_\textrm{account}$ is to enforce authentication on Chatbot Arena through integration with existing digital identity providers. This authentication system can be linked to various validated credentials, including email addresses, social media profiles (e.g., Twitter, Facebook), or phone numbers. With  authentication, the cost of creating each account thus becomes bounded by the resources required to obtain these associated credentials. Risk-based authentication or multi-factor authentication may also be offered through some digital identity providers to increase $c_\textrm{account}$ with limited impact to benign users~\citep{makowski2023evaluation, gavazzi2023study}. Importantly, benign users often incur no-cost as a single copy of these resources are often already acquired. This mitigation may, however, result in distributional shifts as users may engage with Chatbot Arena differently once assumptions of anonymity are removed \citep{chui2014multi}.

\subsubsection{Rate Limiting}
\label{subsubsec:rate}

Reducing $m$ through temporal rate limits on actions for each account is also an effective strategy. Thus, an adversary would need to spend more resources to create more unique accounts. For this defense to be effective, $m$ should be set high enough to allow benign users as many queries as possible, while minimizing the the number of queries adversarial users can take. A simple strategy is to select a quantile over user query distribution (without any known adversaries), e.g., the median. With estimates for the benign query distribution, the choice in $m$ can be refined.

\subsubsection{Malicious User Identification}
\label{subsubsec:user_detection}

Risk-based authentication~\citep{gavazzi2023study} in general leverages user behavior patterns to identify malicious users and increase their action costs. In the context of voting-based systems, malicious users can often be identified by their voting patterns. Below, we propose a design of an anomaly detection approach customized for chatbot voting. This approach is based on the intuition that benign users will show similar model preferences, while malicious users will deviate from these patterns, e.g., by voting for specific models more often. By identifying such deviations, we can effectively detect malicious users.

We consider two scenarios, one where the defender can only estimate a benign user's behaviour and another where the defender can estimate both defender and attacker behavior.

\textbf{Scenario 1: Known Benign Distribution}

In this scenario, we assume that a defender can estimate the expected behaviour for benign users using historical data from previous votes. Now, if an adversary behaves significantly differently from the expected behaviour, the defender can detect it. To do so, we use a likelihood test to differentiate between the null hypothesis $H_\textrm{benign}$ that the user's voting pattern matches the known benign distribution or the alternative hypothesis $H_{\neg \textrm{benign}}$ that the user is from a different source.

Let $x = (x_1, ..., x_n)$ represent a sequence of observed impressions by a user, where each $x_i$ is an impression for one of the available models. Under the null hypothesis $H_\textrm{benign}$, we assume these votes come from the known benign user profile. Also we assume each vote is independent of each other. 

The likelihood of observing the entire sequence under the null hypothesis is then:
\begin{equation}
L(x|H_\textrm{benign}) = \prod_{i=1}^{n} \Pr(x_i|H_\textrm{benign}).
\end{equation}
To assess how extreme this observation is under the null hypothesis, we use the test statistic:
\begin{equation}
T(x) = -2\ln(L(x|H_\textrm{benign})).
\end{equation}
To determine statistical significance, we simulate $m$ sequences under the null hypothesis, where each vote is generated according to the known benign probabilities. For each simulated sequence $s^j$, we calculate its test statistic $T(s^j)$. The empirical p-value is then computed as:
\begin{equation}
p = \frac{1}{m}\sum_{j=1}^{m} I\{T(s^j) \geq T(x)\}
\end{equation}
where $I\{\}$ is the indicator function. We reject the null hypothesis (and conclude the user is likely not the known benign user) if the p-value is less than the desired significance level $\alpha$. In particular we use $\alpha=0.01$ in our evaluations.

\textbf{Scenario 2: Known Benign and Malicious Distributions}

Because the leaderboard is public, the adversary can use the published ratings and counts to make themselves more difficult to detect by mimicking the average user behavior. To this end, the defender can instead release perturbed rankings and counts to each user so as to reduce an attacker's knowledge of the true values. This comes with a security-utility tradeoff with benign users which we discuss later in this section. 

We use the same null hypothesis $H_\textrm{benign}$ and alternative hypothesis $H_{\neg \textrm{benign}}$. Similarly, let $\Pr_B(i),i\in[n]$ be the probability of a benign user voting for model $i$ and $\Pr_{\neg B}(i)$ the same for adversarial users. However, note that  $\Pr_{\neg B}(i)$ will match the perturbed votes released by the defender. We can use the Neyman-Pearson Lemma to construct the hypothesis test. The Neyman-Pearson Lemma states that the optimal decision rule is based on the likelihood ratio.

The likelihood ratio is defined as:
\begin{equation}
    \Lambda(x) = \frac{\Pr_M(x)}{\Pr_B(x)}
\end{equation}

The Bradley-Terry coefficient rating difference between two models defines the probability with which one will be preferred over the other. We can use this to calculate the entire probability distribution $\Pr_B(i)$ and $\Pr_{\neg B}(i)$. Given two models $i$ and $j$ with ratings $Q_i$ and $Q_j$ respectively, the probability that $i$ is preferred is typically modeled using a logistic function as:
\begin{equation}
\Pr(i \text{ preferred over } j) = \frac{1}{1 + \exp(-(Q_i - Q_j) / s)}
\end{equation}

where $s$ is a scaling factor that determines the sensitivity of the probability to the rating difference. Then, we can calculate any component $\Pr_{B}(i)$ (or $\Pr_{\neg B}(i)$ similarly) as the event that this model is chosen over each other model. This is calculated as:
\begin{equation}
    \Pr_B(i)=\prod_j \Pr_B(i \text{ preferred over } j \mid \text{true Bradley-Terry coefficient ratings})
\end{equation}
For $\Pr_{\neg B}(i)$, the perturbed Bradley-Terry coefficient rankings are used instead.

\subsubsection{Increasing $c_\textrm{action}$}
\label{subsubsec:action}

Alternatively, the defender can implement additional security measures to increase the cost of each action an attacker must perform. We list two possible mitigations:
\begin{itemize}[leftmargin=16pt]
    \item Requiring a CAPTCHA per impression/vote: this makes the cost $c_\textrm{action} = N \times c_\textrm{CAPTCHA}$, since automated CAPTCHA-solving services typically charge on a per-CAPTCHA basis.
    
    \item Enforcing prompt uniqueness: A potentially more effective mitigation is to reject or down-weight previously used prompts when updating the Bradley-Terry coefficient leaderboard. This forces attackers to generate new prompts and train corresponding detectors for each action. As detailed in \cref{subsec:detector}, this approach would introduce a cost of approximately \$2.20 per prompt (or per action). However, this mitigation may be ineffective for naturally identifiable models, such as those with output watermarks that the attacker can detect, as discussed in \cref{sec:discussion}.
\end{itemize}

\subsection{Experiments}
Preventing a well resourced adversary in the limit would be almost unfeasible since the adversary could hire many users to submit legitimate votes and avoid any detection. Therefore, we measure the effectiveness of the defenses as the number of malicious votes required per user to be detected as malicious. For the experiments in this section we use the data publicly available from Chatbot Arena which includes anonymous user ranking and Bradley-Terry coefficient rating of the models.

\begin{figure}[t]
  \centering
  \begin{minipage}{0.3\textwidth}
  
    \centering
    
    \includegraphics[width=\textwidth]{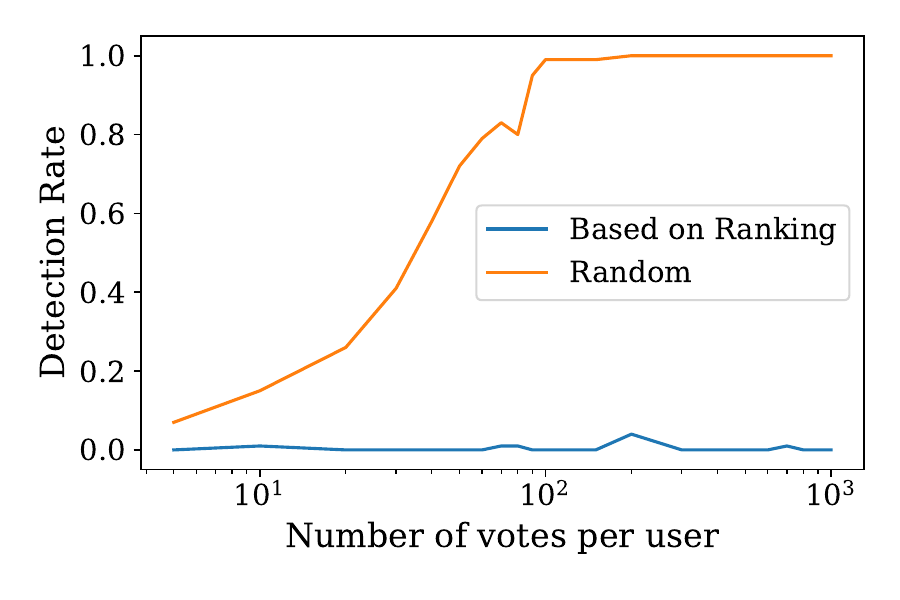} 
    \caption{Scenario 1: The defender uses the likelihood to identify the malicious users. For a naive adversary who randomly chooses between untargetted models this approach can be effective, however, if the adversary uses existing public ranking it can bypass detection }\label{fig:def_scen1}
  \end{minipage}
  \hfill
  \begin{minipage}{0.3\textwidth}
    \centering
      
    \includegraphics[width=\textwidth]{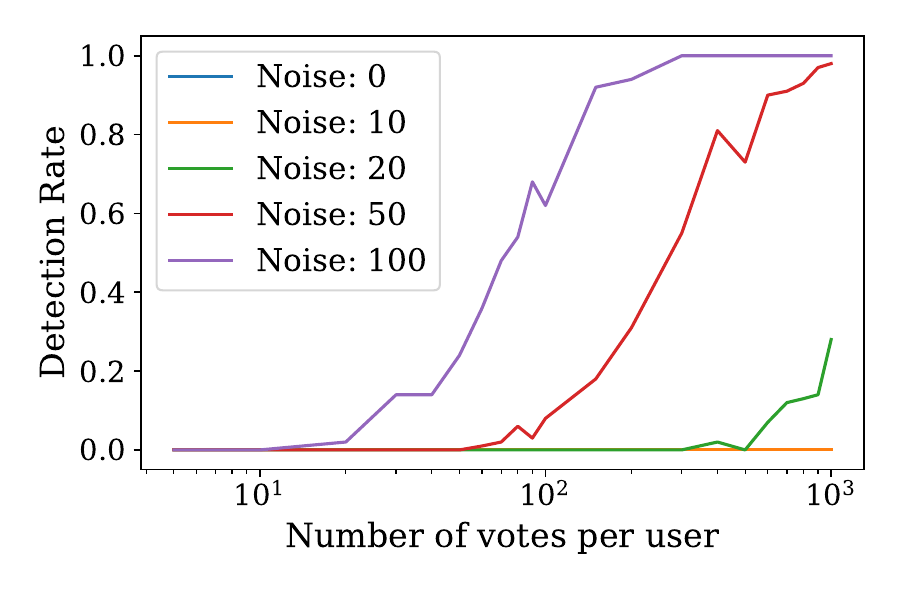}
    \caption{Scenario 2: The defender releases a perturbed version of the leaderboard. Even when an adversary uses this perturbed leaderboard to choose between two untargeted models, their actions can still be detected. Increasing the amount of noise helps in detecting malicious users.}\label{fig:def_scen2}
  \end{minipage}
  \hfill
\begin{minipage}{0.3\textwidth}
    \centering
      
    \includegraphics[width=\textwidth]{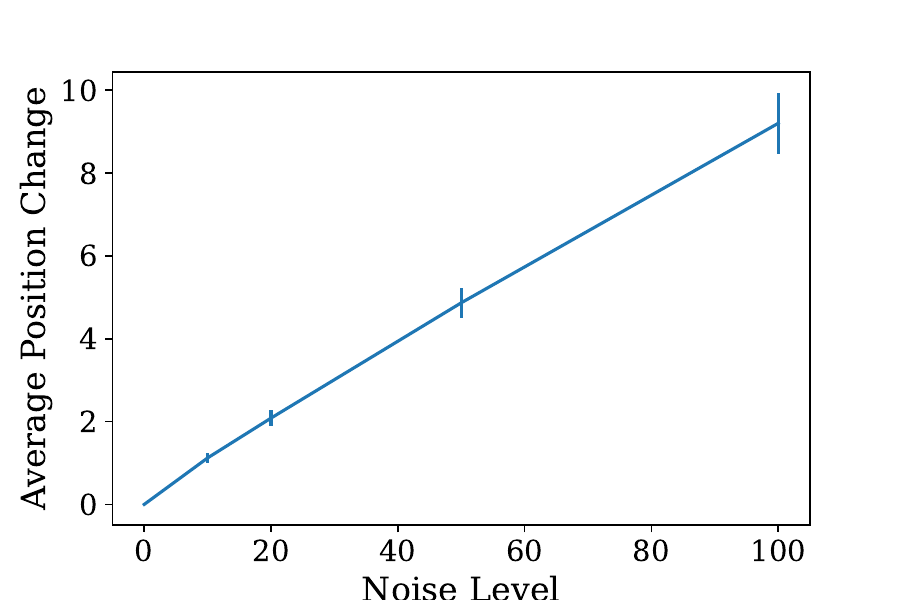}
    \caption{Larger noises significantly change the order of rank list\newline\newline\newline\newline\newline\newline\newline}
    \label{fig:noise_effect}
\end{minipage}
\end{figure}

We start with the first scenario where the defender has access to historical data of the votes between users and can use them to estimate the preferences of a benign user between two models. Figure~\ref{fig:def_scen1} illustrated the results. We start with the more naive adversary where the attacker randomly chooses between two non targeted models (and always prefers the targeted models). As can be seen in the results, the defender can use the difference in the behavior of a random adversary to identify the malicious users. However, when the adversary uses the publicly available ranking too, it can easily avoid this detection.

In the second scenario the defender modifies the rating of the model and releases the perturbed leaderboard. Now if the adversary uses this perturbed order, its behavior can be detected. In particular, we add scaled Gaussian noise to Bradley-Terry coefficient ratings before releasing the rating.  Figures ~\ref{fig:def_scen2} and ~\ref{fig:noise_effect} show the effectiveness and also utility effect of this mitigation approach. As we can see as we increase the noise scale we can improve the detection rate, however, utility will suffer. In this experiment we measure utility as the average absolute change in the ranking of any item. 

As mentioned earlier, while we cannot prevent this attack completely using either authentication approaches or the malicious user detection approach described in this section, we can increase the cost of the attack significantly.  

\section{Related Work}

\textbf{Security vulnerabilities in voting-based system.} 
Voting-based systems are frequently used in security relevant scenarios, such as for malware identification \citep{virustotal_reports} or for content validation \citep{kamvar2003eigentrust}. As a result, attacks on these systems are well studied \citep{hoffman2009survey} and a common approach to securing these systems is to produce reputation scores for users through their voting history \citep{kamvar2003eigentrust, zhai2016anonrep}. We consider an extention of reputation systems to a Chatbot Arena in \cref{subsubsec:account}. In the context of machine learning, reputation has also been used by FLTrust \citep{cao2020fltrust} to defend against \emph{data poisoning} attacks.

\textbf{Detecting the target model for the generation.}
Our primary attack involves training a classifier that can identify which language model system produced a given generation.
This task is related to the much older task of authorship attribution---identifying the authors of anonymous (but human-written) works of writing \citep{huang2024authorship, sun2020deanon}. 
\citet{tay2020reverse} showed how both simple bag-of-words-based classifiers as well as trained neural networks could be used to classify the model configuration used to generate text.
Others have finetuned pre-trained language models such as XLNet \citep{munir2021through} or RoBERTa \citep{wang2024m4gt}, for the task of classifying which pre-trained language model generated a synthetic text sequence.
Our framing of the task is easier than that of most prior work in this space because we assume the attacker has control over the prompt being used for generation, and the set of possible model configurations which may have been used for generation is fairly constrained.

The most related work to ours is the concurrent effort by \cite{zhao2024challengestrustworthyhumanevaluation}, which also investigates the use of targeted model detection algorithms to enable adversarial voting. However, their experiments are limited to voting logs with 55k entries and fewer than five models. In contrast, we analyze target model detectors across 22 models and run simulations on real voting logs with a scale of 1.7 million votes. Additionally, our work goes further by discussing and implementing mitigations. 

\textbf{Evaluation of LLMs.} Various benchmarks have been developed, ranging from general tasks~\citep{hendrycks2021measuring, zellers2019hellaswag,srivastava2023beyond} to specialized domains like math~\citep{cobbe2021training, hendrycks2021measuring}, coding~\citep{chen2021evaluating,austin2021program}, knowledge-intensive applications~\citep{rein2023gpqa}, specific language capabilities like reading comprehension~\citep{dua2019drop} and multilinguality~\citep{shi2023language,lai2023okapi}. However, there are many challenges when using those benchmarks to track the progress of model developments: 1) academic benchmarks focus on measuring fundamental capabilities, which do not always correlate well with application scenarios that average real world users care about~\citep{kopf2024openassistant,zheng2023judging,zheng2023lmsys}; 2) faithfully evaluating open-ended responses to complex questions (e.g. summarization) is highly non-trivial, and it is challenging to quantify the reliability and robustness of current metrics based either on text matching derived heuristics~\citep{liu2008correlation,cohan2016revisiting,fabbri2021summeval} or auto-evaluation with a rating LLM~\citep{zheng2023judging,kim2023prometheus,zhu2023judgelm,wu2024conceptmix, xie2024memorization}; 3) publicly released benchmarks have high risk of data contamination, leading to potentially inaccurate evaluation results~\citep{magar2022data, balloccu2024leak, shidetecting, xu2024benchmarking, oren2023proving}. As a results, evaluation results based on human voting are considered highly valuable signals by all major model developers as it reflects real world user queries and preferences --- the Chatbot Arena leaderboard currently hosts 157 models from more than 20 different model developers. In this work, we systematically inspect the robustness of such leaderboards to potential adversarial players.
\section{Discussion}
\label{sec:discussion}

\textbf{Upvoting one's own models vs downvoting those of a competitor.} It is far easier for a model owner to upvote their own model(s) than to downvote (or upvote) another. Model owners have much more knowledge about their models. They know the entire training dataset and can evaluate the loss on each sample to determine the easiest samples to detect. Further, if their model is deployed as an API, they could simply log  generations that the API produces, and then check each candidate in Chatbot Arena against this database. Finally, the model owner can also strategically make text more detectable, either by using stealthy watermarks that only they have direct knowledge of or by using hidden backdoors on specific prompts. In contrast, our approach in  \Cref{subsec:detect_methods} aims to address the case where the adversary does not necessarily have control over the models whose scores they aim to manipulate.

\textbf{Detection via watermarking.}
There has been a slew of recent research aiming to watermark generated text to identify whether given text was generated with a particular, watermarked model~\citep{kirchenbauer23watermark,kuditipudi2024robust,christ24undetectable}. 
This is indeed a way of breaking model anonymity but it has limited applicability for our task.
Not all models employ watermarking, and successful de-anonymization would require the attacker to know the specifics of the watermarking implementation in the target models---information that is typically not public.

\textbf{Implications for public evaluation of AI systems.}
While this paper focuses on Chatbot Arena, our findings our relevant for any public platform for performing comparative evaluation of AI systems, such as ones deployed for evaluating text-to-image and speech.\footnote{\url{https://artificialanalysis.ai}} There is a fundamental tension when designing human evaluation experiments. On one hand, human evaluation paradigms that closely reflect real-world usage lend validity to the results. On the other hand, restricting human evaluation to known groups of annotators lends greater control annotator qualifications, demographic makeup, and incentives---but at the expense of the transferability of the findings to real-world usage. For example, prior work has shown that Amazon Mechanical Turk workers rate generated text very differently than school teachers \citep{karpinska2021perils}.
\section{Conclusions}
\label{sec:conclusions}

The field of natural language processing has long relied on domain-specific, easy-to-implement evaluation metrics. But dramatic advances in LLM performance challenges traditional evaluation practices. 
%
As we show in this paper, moving from evaluations that use an objective source of truth to evaluations that utilize human inputs introduces the potential for new types of evaluation difficulties.
We focus on this paper in validating one straightforward attack: by identifying and selectively voting for (or against) a particular model, an adversary can significantly alter the ordering of the best models.

Mitigating this attack is feasible, and we are actively collaborating with the Chatbot Arena team to make Chatbot Arena more robust. 
We also encourage the community to explore and adopt mitigation strategies, such as voter authentication, rate limits, and more robust mechanisms for detecting malicious activities.

More broadly, however, the shift from \emph{objective} to \emph{subjective} language model evaluations opens the potential for new forms of evaluation failures.
Our paper explores just one of these failure modes---where an adversary explicitly aims to alter the rank of a particular target model.
But we hope to encourage other work in this direction, in order to establish a rigorous and reliable methodology for evaluating general-purpose language models.

\section*{Ethics and Disclosure}

Our study highlights the susceptibility of Chatbot Arena's leaderboard rankings to malicious voting behavior. We conducted this work with the goal of improving the security and reliability of interactive evaluation platforms, and to encourage the development of countermeasures to improve robustness.

We disclosed this attack in August 2024 and collaborated with the Chatbot Arena team throughout the development of this work to assist in developing appropriate defenses. Our collaboration has been instrumental in refining solutions to mitigate these vulnerabilities, ensuring that platform integrity and user trust are maintained. By sharing these results, we aim to encourage the community to adopt stronger safeguards in the design and evaluation of similar systems.

All simulations and experiments conducted in this study were carried out in a controlled environment, with no real-world impact on the existing Chatbot Arena platform or any other public-facing system.

Finally, as concurrent work has begun to raise similar issues in voting-based ranking systems~\citep{zhao2024challengestrustworthyhumanevaluation}, we believe there is little marginal increase in risk from releasing our paper.

\section*{Contribution Statement}

This project was a team effort.
\begin{itemize}[nosep, leftmargin=16pt]
    \item \textbf{Idea formulation:}  Yangsibo came up with the idea of using model de-identification to manipulate leaderboard rankings on Chatbot Arena. Nicholas and Florian suggested running simulations to quantify the attack efficacy via estimating the number of votes required to shift models' positions on the leaderboard. 
    \item \textbf{De-anonymizing models} (\cref{sec:detect}):  Milad suggested using $\mathsf{TF\mathrm{-}IDF}$ and $\mathsf{BoW}$ for training-based detectors, and Yangsibo conducted experiments demonstrating their effectiveness. Ken suggested and explored the identity-probing detector. Yangsibo collaborated with Ken to finalize results.
    \item \textbf{Disclosure with Chatbot Arena:} In August 2024, Yangsibo, Milad, Chiyuan, and Nicholas contacted the Chatbot Arena team (Wei-Lin, Anastasios, and Ion) to disclose their findings that anonymous model responses can be de-identified with very high accuracy. The Chatbot Arena team expressed interest in collaborating to investigate and address this security vulnerability. As a result, Yangsibo, Milad, Nicholas, Chiyuan, Wei-Lin, Anastasios, and Ion began having regular meetings to advance the project.
    \item \textbf{Estimating number of adversarial votes} (\cref{sec:vote}): The Chatbot Arena team shared a simulation platform. Yangsibo conducted the simulations to estimate the number of votes needed by the attack, with feedback from Milad, Chiyuan, Nicholas and the Chatbot Arena team.
    \item \textbf{Exploring mitigations} (\cref{sec:defense}): Ion suggested exploring mitigation strategies. Milad, Yangsibo, Chiyuan, and Nicholas developed the attack cost model (\cref{subsec:cost}) and refined it with input from the Chatbot Arena team. For mitigations, Nicholas suggested authentication (\cref{subsubsec:auth}) and rate limiting (\cref{subsubsec:rate}); Anastasios suggested exploring customized malicious user identification algorithms, and then Milad drafted the proposals with Chris (\cref{subsubsec:user_detection}) and ran experiments; Chiyuan suggested increasing the cost of actions (\cref{subsubsec:action}).
    \item \textbf{Writing:} Yangsibo and Milad prepared the initial draft. Chris, Chiyuan, Katherine, Daphne, Nicholas, Florian, Matthew, Ken, Wei-Lin, Anastasios, and Ion wrote and edited the paper.
    \item \textbf{Paper release:} Katherine, Milad, Chiyuan, and Yangsibo prepared the paper for public release.
\end{itemize}

\section*{Acknowledgments}
We thank Szymon Tworkowski, Samuel Bowman, Zheng-Xin Yong, Mengzhou Xia, Haochen Zhang, Tianle Cai, and Danqi Chen for their valuable discussions during the early stages of this paper. We are grateful to Andreas Terzis, Martin Abadi, Four Flynn, Shira McNamara, Jon Small, Anand Rao, and Aneesh Pappu for comments and reviews.

\clearpage
\bibliography{ref}

\begin{thebibliography}{57}
\providecommand{\natexlab}[1]{#1}
\providecommand{\url}[1]{\texttt{#1}}
\expandafter\ifx\csname urlstyle\endcsname\relax
  \providecommand{\doi}[1]{doi: #1}\else
  \providecommand{\doi}{doi: \begingroup \urlstyle{rm}\Url}\fi

\bibitem[Achiam et~al.(2023)Achiam, Adler, Agarwal, Ahmad, Akkaya, Aleman, Almeida, Altenschmidt, Altman, Anadkat, et~al.]{achiam2023gpt}
Josh Achiam, Steven Adler, Sandhini Agarwal, Lama Ahmad, Ilge Akkaya, Florencia~Leoni Aleman, Diogo Almeida, Janko Altenschmidt, Sam Altman, Shyamal Anadkat, et~al.
\newblock Gpt-4 technical report.
\newblock \emph{arXiv preprint arXiv:2303.08774}, 2023.

\bibitem[{Anthropic}(2024)]{Anthropic2024}
{Anthropic}.
\newblock Anthropic introduces the claude 3 model family, March 2024.
\newblock Available at: \url{https://www.anthropic.com/news/claude-3-family}.

\bibitem[Austin et~al.(2021)Austin, Odena, Nye, Bosma, Michalewski, Dohan, Jiang, Cai, Terry, Le, et~al.]{austin2021program}
Jacob Austin, Augustus Odena, Maxwell Nye, Maarten Bosma, Henryk Michalewski, David Dohan, Ellen Jiang, Carrie Cai, Michael Terry, Quoc Le, et~al.
\newblock Program synthesis with large language models.
\newblock \emph{arXiv preprint arXiv:2108.07732}, 2021.

\bibitem[Balloccu et~al.(2024)Balloccu, Schmidtov{\'a}, Lango, and Dusek]{balloccu2024leak}
Simone Balloccu, Patr{\'\i}cia Schmidtov{\'a}, Mateusz Lango, and Ondrej Dusek.
\newblock Leak, cheat, repeat: Data contamination and evaluation malpractices in closed-source {LLM}s.
\newblock In Yvette Graham and Matthew Purver (eds.), \emph{Proceedings of the 18th Conference of the European Chapter of the Association for Computational Linguistics (Volume 1: Long Papers)}, pp.\  67--93, St. Julian{'}s, Malta, March 2024. Association for Computational Linguistics.
\newblock URL \url{https://aclanthology.org/2024.eacl-long.5}.

\bibitem[Cao et~al.(2020)Cao, Fang, Liu, and Gong]{cao2020fltrust}
Xiaoyu Cao, Minghong Fang, Jia Liu, and Neil~Zhenqiang Gong.
\newblock Fltrust: Byzantine-robust federated learning via trust bootstrapping.
\newblock \emph{arXiv preprint arXiv:2012.13995}, 2020.

\bibitem[Chen et~al.(2021)Chen, Tworek, Jun, Yuan, Pinto, Kaplan, Edwards, Burda, Joseph, Brockman, et~al.]{chen2021evaluating}
Mark Chen, Jerry Tworek, Heewoo Jun, Qiming Yuan, Henrique Ponde De~Oliveira Pinto, Jared Kaplan, Harri Edwards, Yuri Burda, Nicholas Joseph, Greg Brockman, et~al.
\newblock Evaluating large language models trained on code.
\newblock \emph{arXiv preprint arXiv:2107.03374}, 2021.

\bibitem[Chiang et~al.(2024)Chiang, Zheng, Sheng, Angelopoulos, Li, Li, Zhu, Zhang, Jordan, Gonzalez, and Stoica]{chiang2024chatbot}
Wei-Lin Chiang, Lianmin Zheng, Ying Sheng, Anastasios~Nikolas Angelopoulos, Tianle Li, Dacheng Li, Banghua Zhu, Hao Zhang, Michael Jordan, Joseph~E. Gonzalez, and Ion Stoica.
\newblock Chatbot arena: An open platform for evaluating {LLM}s by human preference.
\newblock In \emph{Forty-first International Conference on Machine Learning}, 2024.
\newblock URL \url{https://openreview.net/forum?id=3MW8GKNyzI}.

\bibitem[Christ et~al.(2024)Christ, Gunn, and Zamir]{christ24undetectable}
Miranda Christ, Sam Gunn, and Or~Zamir.
\newblock Undetectable watermarks for language models.
\newblock In Shipra Agrawal and Aaron Roth (eds.), \emph{Proceedings of Thirty Seventh Conference on Learning Theory}, volume 247 of \emph{Proceedings of Machine Learning Research}, pp.\  1125--1139. PMLR, 30 Jun--03 Jul 2024.
\newblock URL \url{https://proceedings.mlr.press/v247/christ24a.html}.

\bibitem[Chui(2014)]{chui2014multi}
Rebecca Chui.
\newblock A multi-faceted approach to anonymity online: Examining the relations between anonymity and antisocial behaviour.
\newblock \emph{Journal For Virtual Worlds Research}, 7\penalty0 (2), 2014.

\bibitem[Cobbe et~al.(2021)Cobbe, Kosaraju, Bavarian, Chen, Jun, Kaiser, Plappert, Tworek, Hilton, Nakano, et~al.]{cobbe2021training}
Karl Cobbe, Vineet Kosaraju, Mohammad Bavarian, Mark Chen, Heewoo Jun, Lukasz Kaiser, Matthias Plappert, Jerry Tworek, Jacob Hilton, Reiichiro Nakano, et~al.
\newblock Training verifiers to solve math word problems.
\newblock \emph{arXiv preprint arXiv:2110.14168}, 2021.

\bibitem[Cohan \& Goharian(2016)Cohan and Goharian]{cohan2016revisiting}
Arman Cohan and Nazli Goharian.
\newblock Revisiting summarization evaluation for scientific articles.
\newblock In Nicoletta Calzolari, Khalid Choukri, Thierry Declerck, Sara Goggi, Marko Grobelnik, Bente Maegaard, Joseph Mariani, Helene Mazo, Asuncion Moreno, Jan Odijk, and Stelios Piperidis (eds.), \emph{Proceedings of the Tenth International Conference on Language Resources and Evaluation ({LREC}'16)}, pp.\  806--813, Portoro{\v{z}}, Slovenia, May 2016. European Language Resources Association (ELRA).
\newblock URL \url{https://aclanthology.org/L16-1130}.

\bibitem[Dua et~al.(2019)Dua, Wang, Dasigi, Stanovsky, Singh, and Gardner]{dua2019drop}
Dheeru Dua, Yizhong Wang, Pradeep Dasigi, Gabriel Stanovsky, Sameer Singh, and Matt Gardner.
\newblock Drop: A reading comprehension benchmark requiring discrete reasoning over paragraphs.
\newblock In \emph{Proceedings of the 2019 Conference of the North American Chapter of the Association for Computational Linguistics: Human Language Technologies, Volume 1 (Long and Short Papers)}, pp.\  2368--2378, 2019.

\bibitem[Dubey et~al.(2024)Dubey, Jauhri, Pandey, Kadian, Al-Dahle, Letman, Mathur, Schelten, Yang, Fan, et~al.]{dubey2024llama}
Abhimanyu Dubey, Abhinav Jauhri, Abhinav Pandey, Abhishek Kadian, Ahmad Al-Dahle, Aiesha Letman, Akhil Mathur, Alan Schelten, Amy Yang, Angela Fan, et~al.
\newblock The llama 3 herd of models.
\newblock \emph{arXiv preprint arXiv:2407.21783}, 2024.

\bibitem[Fabbri et~al.(2021)Fabbri, Kry{\'s}ci{\'n}ski, McCann, Xiong, Socher, and Radev]{fabbri2021summeval}
Alexander~R Fabbri, Wojciech Kry{\'s}ci{\'n}ski, Bryan McCann, Caiming Xiong, Richard Socher, and Dragomir Radev.
\newblock Summeval: Re-evaluating summarization evaluation.
\newblock \emph{Transactions of the Association for Computational Linguistics}, 9:\penalty0 391--409, 2021.

\bibitem[Gavazzi et~al.(2023)Gavazzi, Williams, Kirda, Lu, King, Davis, and Leek]{gavazzi2023study}
Anthony Gavazzi, Ryan Williams, Engin Kirda, Long Lu, Andre King, Andy Davis, and Tim Leek.
\newblock A study of $\{$Multi-Factor$\}$ and $\{$Risk-Based$\}$ authentication availability.
\newblock In \emph{32nd USENIX Security Symposium (USENIX Security 23)}, pp.\  2043--2060, 2023.

\bibitem[Hendrycks et~al.(2021)Hendrycks, Burns, Basart, Zou, Mazeika, Song, and Steinhardt]{hendrycks2021measuring}
Dan Hendrycks, Collin Burns, Steven Basart, Andy Zou, Mantas Mazeika, Dawn Song, and Jacob Steinhardt.
\newblock Measuring massive multitask language understanding.
\newblock In \emph{International Conference on Learning Representations}, 2021.
\newblock URL \url{https://openreview.net/forum?id=d7KBjmI3GmQ}.

\bibitem[Hoffman et~al.(2009)Hoffman, Zage, and Nita-Rotaru]{hoffman2009survey}
Kevin Hoffman, David Zage, and Cristina Nita-Rotaru.
\newblock A survey of attack and defense techniques for reputation systems.
\newblock \emph{ACM Computing Surveys (CSUR)}, 42\penalty0 (1):\penalty0 1--31, 2009.

\bibitem[Huang et~al.(2024)Huang, Chen, and Shu]{huang2024authorship}
Baixiang Huang, Canyu Chen, and Kai Shu.
\newblock Authorship attribution in the era of llms: Problems, methodologies, and challenges.
\newblock \emph{arXiv preprint arXiv:2408.08946}, 2024.

\bibitem[Hunter(2004)]{hunter2004mm}
David~R Hunter.
\newblock Mm algorithms for generalized bradley-terry models.
\newblock \emph{The annals of statistics}, 32\penalty0 (1):\penalty0 384--406, 2004.

\bibitem[Kamvar et~al.(2003)Kamvar, Schlosser, and Garcia-Molina]{kamvar2003eigentrust}
Sepandar~D Kamvar, Mario~T Schlosser, and Hector Garcia-Molina.
\newblock The eigentrust algorithm for reputation management in p2p networks.
\newblock In \emph{Proceedings of the 12th international conference on World Wide Web}, pp.\  640--651, 2003.

\bibitem[Karpinska et~al.(2021)Karpinska, Akoury, and Iyyer]{karpinska2021perils}
Marzena Karpinska, Nader Akoury, and Mohit Iyyer.
\newblock The perils of using {M}echanical {T}urk to evaluate open-ended text generation.
\newblock In Marie-Francine Moens, Xuanjing Huang, Lucia Specia, and Scott Wen-tau Yih (eds.), \emph{Proceedings of the 2021 Conference on Empirical Methods in Natural Language Processing}, pp.\  1265--1285, Online and Punta Cana, Dominican Republic, November 2021. Association for Computational Linguistics.
\newblock \doi{10.18653/v1/2021.emnlp-main.97}.
\newblock URL \url{https://aclanthology.org/2021.emnlp-main.97}.

\bibitem[Kim et~al.(2023)Kim, Shin, Cho, Jang, Longpre, Lee, Yun, Shin, Kim, Thorne, et~al.]{kim2023prometheus}
Seungone Kim, Jamin Shin, Yejin Cho, Joel Jang, Shayne Longpre, Hwaran Lee, Sangdoo Yun, Seongjin Shin, Sungdong Kim, James Thorne, et~al.
\newblock Prometheus: Inducing fine-grained evaluation capability in language models.
\newblock In \emph{The Twelfth International Conference on Learning Representations}, 2023.

\bibitem[Kirchenbauer et~al.(2023)Kirchenbauer, Geiping, Wen, Katz, Miers, and Goldstein]{kirchenbauer23watermark}
John Kirchenbauer, Jonas Geiping, Yuxin Wen, Jonathan Katz, Ian Miers, and Tom Goldstein.
\newblock A watermark for large language models.
\newblock In Andreas Krause, Emma Brunskill, Kyunghyun Cho, Barbara Engelhardt, Sivan Sabato, and Jonathan Scarlett (eds.), \emph{Proceedings of the 40th International Conference on Machine Learning}, volume 202 of \emph{Proceedings of Machine Learning Research}, pp.\  17061--17084. PMLR, 23--29 Jul 2023.
\newblock URL \url{https://proceedings.mlr.press/v202/kirchenbauer23a.html}.

\bibitem[K{\"o}pf et~al.(2024)K{\"o}pf, Kilcher, von R{\"u}tte, Anagnostidis, Tam, Stevens, Barhoum, Nguyen, Stanley, Nagyfi, et~al.]{kopf2024openassistant}
Andreas K{\"o}pf, Yannic Kilcher, Dimitri von R{\"u}tte, Sotiris Anagnostidis, Zhi~Rui Tam, Keith Stevens, Abdullah Barhoum, Duc Nguyen, Oliver Stanley, Rich{\'a}rd Nagyfi, et~al.
\newblock Openassistant conversations-democratizing large language model alignment.
\newblock \emph{Advances in Neural Information Processing Systems}, 36, 2024.

\bibitem[Kuditipudi et~al.(2024)Kuditipudi, Thickstun, Hashimoto, and Liang]{kuditipudi2024robust}
Rohith Kuditipudi, John Thickstun, Tatsunori Hashimoto, and Percy Liang.
\newblock Robust distortion-free watermarks for language models.
\newblock \emph{Transactions on Machine Learning Research}, 2024.
\newblock ISSN 2835-8856.
\newblock URL \url{https://openreview.net/forum?id=FpaCL1MO2C}.

\bibitem[Lai et~al.(2023)Lai, Nguyen, Ngo, Nguyen, Dernoncourt, Rossi, and Nguyen]{lai2023okapi}
Viet Lai, Chien Nguyen, Nghia Ngo, Thuat Nguyen, Franck Dernoncourt, Ryan Rossi, and Thien Nguyen.
\newblock Okapi: Instruction-tuned large language models in multiple languages with reinforcement learning from human feedback.
\newblock In Yansong Feng and Els Lefever (eds.), \emph{Proceedings of the 2023 Conference on Empirical Methods in Natural Language Processing: System Demonstrations}, pp.\  318--327, Singapore, December 2023. Association for Computational Linguistics.
\newblock \doi{10.18653/v1/2023.emnlp-demo.28}.
\newblock URL \url{https://aclanthology.org/2023.emnlp-demo.28}.

\bibitem[Lassak et~al.(2024)Lassak, Pan, Ur, and Golla]{lassak2024aren}
Leona Lassak, Elleen Pan, Blase Ur, and Maximilian Golla.
\newblock Why aren{\textquoteright}t we using passkeys? obstacles companies face deploying {FIDO2} passwordless authentication.
\newblock In \emph{33rd USENIX Security Symposium (USENIX Security 24)}, pp.\  7231--7248, Philadelphia, PA, August 2024. USENIX Association.
\newblock ISBN 978-1-939133-44-1.
\newblock URL \url{https://www.usenix.org/conference/usenixsecurity24/presentation/lassak}.

\bibitem[Li et~al.(2024)Li, Held, Ryan, Pipatanakul, Manakul, Zhu, and Yang]{talkarena2024}
Minzhi Li, Will Held, Michael~J. Ryan, Kunat Pipatanakul, Potsawee Manakul, Hao Zhu, and Diyi Yang.
\newblock Talk arena: Interactive evaluation of large audio models, 2024.

\bibitem[Liu \& Liu(2008)Liu and Liu]{liu2008correlation}
Feifan Liu and Yang Liu.
\newblock Correlation between rouge and human evaluation of extractive meeting summaries.
\newblock In \emph{Proceedings of ACL-08: HLT, short papers}, pp.\  201--204, 2008.

\bibitem[Lu et~al.(2024)Lu, Jiang, Chen, Wang, Choi, and Lin]{lu2024wildvision}
Yujie Lu, Dongfu Jiang, Wenhu Chen, William~Yang Wang, Yejin Choi, and Bill~Yuchen Lin.
\newblock Wildvision: Evaluating vision-language models in the wild with human preferences.
\newblock \emph{arXiv preprint arXiv:2406.11069}, 2024.

\bibitem[Magar \& Schwartz(2022)Magar and Schwartz]{magar2022data}
Inbal Magar and Roy Schwartz.
\newblock Data contamination: From memorization to exploitation.
\newblock In Smaranda Muresan, Preslav Nakov, and Aline Villavicencio (eds.), \emph{Proceedings of the 60th Annual Meeting of the Association for Computational Linguistics (Volume 2: Short Papers)}, pp.\  157--165, Dublin, Ireland, May 2022. Association for Computational Linguistics.
\newblock \doi{10.18653/v1/2022.acl-short.18}.
\newblock URL \url{https://aclanthology.org/2022.acl-short.18}.

\bibitem[Makowski \& P{\"o}hn(2023)Makowski and P{\"o}hn]{makowski2023evaluation}
Jan-Phillip Makowski and Daniela P{\"o}hn.
\newblock Evaluation of real-world risk-based authentication at online services revisited: Complexity wins.
\newblock In \emph{Proceedings of the 18th International Conference on Availability, Reliability and Security}, pp.\  1--9, 2023.

\bibitem[Munir et~al.(2021)Munir, Batool, Shafiq, Srinivasan, and Zaffar]{munir2021through}
Shaoor Munir, Brishna Batool, Zubair Shafiq, Padmini Srinivasan, and Fareed Zaffar.
\newblock Through the looking glass: Learning to attribute synthetic text generated by language models.
\newblock In \emph{Proceedings of the 16th Conference of the European Chapter of the Association for Computational Linguistics: Main Volume}, pp.\  1811--1822, 2021.

\bibitem[Oren et~al.(2024)Oren, Meister, Chatterji, Ladhak, and Hashimoto]{oren2023proving}
Yonatan Oren, Nicole Meister, Niladri Chatterji, Faisal Ladhak, and Tatsunori~B Hashimoto.
\newblock Proving test set contamination in black box language models.
\newblock In \emph{The Twelfth International Conference on Learning Representations}, 2024.
\newblock URL \url{https://openreview.net/forum?id=KS8mIvetg2}.

\bibitem[Reid et~al.(2024)Reid, Savinov, Teplyashin, Lepikhin, Lillicrap, Alayrac, Soricut, Lazaridou, Firat, Schrittwieser, et~al.]{reid2024gemini}
Machel Reid, Nikolay Savinov, Denis Teplyashin, Dmitry Lepikhin, Timothy Lillicrap, Jean-baptiste Alayrac, Radu Soricut, Angeliki Lazaridou, Orhan Firat, Julian Schrittwieser, et~al.
\newblock Gemini 1.5: Unlocking multimodal understanding across millions of tokens of context.
\newblock \emph{arXiv preprint arXiv:2403.05530}, 2024.

\bibitem[Rein et~al.(2023)Rein, Hou, Stickland, Petty, Pang, Dirani, Michael, and Bowman]{rein2023gpqa}
David Rein, Betty~Li Hou, Asa~Cooper Stickland, Jackson Petty, Richard~Yuanzhe Pang, Julien Dirani, Julian Michael, and Samuel~R Bowman.
\newblock Gpqa: A graduate-level google-proof q\&a benchmark.
\newblock \emph{arXiv preprint arXiv:2311.12022}, 2023.

\bibitem[Rumbelow \& Watkins(2023)Rumbelow and Watkins]{solidgoldmagikarp}
Jessica Rumbelow and Matthew Watkins.
\newblock {SolidGoldMagikarp} (plus, prompt generation).
\newblock \url{https://www.lesswrong.com/posts/aPeJE8bSo6rAFoLqg}, 2023.

\bibitem[Salton \& Buckley(1988)Salton and Buckley]{salton1988term}
Gerard Salton and Christopher Buckley.
\newblock Term-weighting approaches in automatic text retrieval.
\newblock \emph{Information processing \& management}, 24\penalty0 (5):\penalty0 513--523, 1988.

\bibitem[Salton et~al.(1975)Salton, Wong, and Yang]{salton1975vector}
Gerard Salton, Anita Wong, and Chung-Shu Yang.
\newblock A vector space model for automatic indexing.
\newblock \emph{Communications of the ACM}, 18\penalty0 (11):\penalty0 613--620, 1975.

\bibitem[Shi et~al.(2023)Shi, Suzgun, Freitag, Wang, Srivats, Vosoughi, Chung, Tay, Ruder, Zhou, Das, and Wei]{shi2023language}
Freda Shi, Mirac Suzgun, Markus Freitag, Xuezhi Wang, Suraj Srivats, Soroush Vosoughi, Hyung~Won Chung, Yi~Tay, Sebastian Ruder, Denny Zhou, Dipanjan Das, and Jason Wei.
\newblock Language models are multilingual chain-of-thought reasoners.
\newblock In \emph{The Eleventh International Conference on Learning Representations}, 2023.
\newblock URL \url{https://openreview.net/forum?id=fR3wGCk-IXp}.

\bibitem[Shi et~al.(2024)Shi, Ajith, Xia, Huang, Liu, Blevins, Chen, and Zettlemoyer]{shidetecting}
Weijia Shi, Anirudh Ajith, Mengzhou Xia, Yangsibo Huang, Daogao Liu, Terra Blevins, Danqi Chen, and Luke Zettlemoyer.
\newblock Detecting pretraining data from large language models.
\newblock In \emph{The Twelfth International Conference on Learning Representations}, 2024.
\newblock URL \url{https://openreview.net/forum?id=zWqr3MQuNs}.

\bibitem[Srivastava et~al.(2023)Srivastava, Rastogi, Rao, Shoeb, Abid, Fisch, Brown, Santoro, Gupta, Garriga-Alonso, , et~al.]{srivastava2023beyond}
Aarohi Srivastava, Abhinav Rastogi, Abhishek Rao, Abu Awal~Md Shoeb, Abubakar Abid, Adam Fisch, Adam~R. Brown, Adam Santoro, Aditya Gupta, Adri{\`a} Garriga-Alonso, , et~al.
\newblock Beyond the imitation game: Quantifying and extrapolating the capabilities of language models.
\newblock \emph{Transactions on Machine Learning Research}, 2023.
\newblock ISSN 2835-8856.
\newblock URL \url{https://openreview.net/forum?id=uyTL5Bvosj}.
\newblock Featured Certification.

\bibitem[Sun et~al.(2020)Sun, Schuster, and Shmatikov]{sun2020deanon}
Zhen Sun, Roei Schuster, and Vitaly Shmatikov.
\newblock De-anonymizing text by fingerprinting language generation.
\newblock In H.~Larochelle, M.~Ranzato, R.~Hadsell, M.F. Balcan, and H.~Lin (eds.), \emph{Advances in Neural Information Processing Systems}, volume~33, pp.\  22420--22431. Curran Associates, Inc., 2020.
\newblock URL \url{https://proceedings.neurips.cc/paper_files/paper/2020/file/fdf2aade29d18910051a6c76ae661860-Paper.pdf}.

\bibitem[Tay et~al.(2020)Tay, Bahri, Zheng, Brunk, Metzler, and Tomkins]{tay2020reverse}
Yi~Tay, Dara Bahri, Che Zheng, Clifford Brunk, Donald Metzler, and Andrew Tomkins.
\newblock Reverse engineering configurations of neural text generation models.
\newblock In \emph{Proceedings of the 58th Annual Meeting of the Association for Computational Linguistics}, pp.\  275--279, 2020.

\bibitem[VirusTotal(2024)]{virustotal_reports}
VirusTotal.
\newblock Results reports.
\newblock VirusTotal Documentation, 2024.
\newblock URL \url{https://docs.virustotal.com/docs/results-reports}.
\newblock Accessed on December 19, 2024.

\bibitem[Wang et~al.(2024)Wang, Mansurov, Ivanov, Su, Shelmanov, Tsvigun, Afzal, Mahmoud, Puccetti, Arnold, et~al.]{wang2024m4gt}
Yuxia Wang, Jonibek Mansurov, Petar Ivanov, Jinyan Su, Artem Shelmanov, Akim Tsvigun, Osama~Mohanned Afzal, Tarek Mahmoud, Giovanni Puccetti, Thomas Arnold, et~al.
\newblock M4gt-bench: Evaluation benchmark for black-box machine-generated text detection.
\newblock \emph{arXiv preprint arXiv:2402.11175}, 2024.

\bibitem[Wu et~al.(2024)Wu, Yu, Huang, Russakovsky, and Arora]{wu2024conceptmix}
Xindi Wu, Dingli Yu, Yangsibo Huang, Olga Russakovsky, and Sanjeev Arora.
\newblock Conceptmix: A compositional image generation benchmark with controllable difficulty.
\newblock \emph{arXiv preprint arXiv:2408.14339}, 2024.

\bibitem[Xie et~al.(2024)Xie, Huang, Zhang, Yu, Chen, Lin, Li, Ghazi, and Kumar]{xie2024memorization}
Chulin Xie, Yangsibo Huang, Chiyuan Zhang, Da~Yu, Xinyun Chen, Bill~Yuchen Lin, Bo~Li, Badih Ghazi, and Ravi Kumar.
\newblock On memorization of large language models in logical reasoning.
\newblock \emph{arXiv preprint arXiv:2410.23123}, 2024.

\bibitem[Xu et~al.(2024)Xu, Wang, Fan, and Liu]{xu2024benchmarking}
Ruijie Xu, Zengzhi Wang, Run-Ze Fan, and Pengfei Liu.
\newblock Benchmarking benchmark leakage in large language models.
\newblock \emph{arXiv preprint arXiv:2404.18824}, 2024.

\bibitem[Zellers et~al.(2019)Zellers, Holtzman, Bisk, Farhadi, and Choi]{zellers2019hellaswag}
Rowan Zellers, Ari Holtzman, Yonatan Bisk, Ali Farhadi, and Yejin Choi.
\newblock {H}ella{S}wag: Can a machine really finish your sentence?
\newblock In Anna Korhonen, David Traum, and Llu{\'\i}s M{\`a}rquez (eds.), \emph{Proceedings of the 57th Annual Meeting of the Association for Computational Linguistics}, pp.\  4791--4800, Florence, Italy, July 2019. Association for Computational Linguistics.
\newblock \doi{10.18653/v1/P19-1472}.
\newblock URL \url{https://aclanthology.org/P19-1472}.

\bibitem[Zhai et~al.(2016)Zhai, Wolinsky, Chen, Syta, Teng, and Ford]{zhai2016anonrep}
Ennan Zhai, David~Isaac Wolinsky, Ruichuan Chen, Ewa Syta, Chao Teng, and Bryan Ford.
\newblock $\{$AnonRep$\}$: Towards $\{$Tracking-Resistant$\}$ anonymous reputation.
\newblock In \emph{13th USENIX Symposium on Networked Systems Design and Implementation (NSDI 16)}, pp.\  583--596, 2016.

\bibitem[Zhao et~al.(2024)Zhao, Rush, and Goyal]{zhao2024challengestrustworthyhumanevaluation}
Wenting Zhao, Alexander~M. Rush, and Tanya Goyal.
\newblock Challenges in trustworthy human evaluation of chatbots, 2024.
\newblock URL \url{https://arxiv.org/abs/2412.04363}.

\bibitem[Zheng et~al.(2023{\natexlab{a}})Zheng, Chiang, Sheng, Li, Zhuang, Wu, Zhuang, Li, Lin, Xing, Gonzalez, Stoica, and Zhang]{zheng2023lmsyschat1m}
Lianmin Zheng, Wei-Lin Chiang, Ying Sheng, Tianle Li, Siyuan Zhuang, Zhanghao Wu, Yonghao Zhuang, Zhuohan Li, Zi~Lin, Eric.~P Xing, Joseph~E. Gonzalez, Ion Stoica, and Hao Zhang.
\newblock Lmsys-chat-1m: A large-scale real-world llm conversation dataset, 2023{\natexlab{a}}.

\bibitem[Zheng et~al.(2023{\natexlab{b}})Zheng, Chiang, Sheng, Li, Zhuang, Wu, Zhuang, Li, Lin, Xing, et~al.]{zheng2023lmsys}
Lianmin Zheng, Wei-Lin Chiang, Ying Sheng, Tianle Li, Siyuan Zhuang, Zhanghao Wu, Yonghao Zhuang, Zhuohan Li, Zi~Lin, Eric~P Xing, et~al.
\newblock Lmsys-chat-1m: A large-scale real-world llm conversation dataset.
\newblock \emph{arXiv preprint arXiv:2309.11998}, 2023{\natexlab{b}}.

\bibitem[Zheng et~al.(2023{\natexlab{c}})Zheng, Chiang, Sheng, Zhuang, Wu, Zhuang, Lin, Li, Li, Xing, et~al.]{zheng2023judging}
Lianmin Zheng, Wei-Lin Chiang, Ying Sheng, Siyuan Zhuang, Zhanghao Wu, Yonghao Zhuang, Zi~Lin, Zhuohan Li, Dacheng Li, Eric Xing, et~al.
\newblock Judging llm-as-a-judge with mt-bench and chatbot arena.
\newblock \emph{Advances in Neural Information Processing Systems}, 36:\penalty0 46595--46623, 2023{\natexlab{c}}.

\bibitem[Zhu et~al.(2023)Zhu, Wang, and Wang]{zhu2023judgelm}
Lianghui Zhu, Xinggang Wang, and Xinlong Wang.
\newblock Judgelm: Fine-tuned large language models are scalable judges.
\newblock \emph{arXiv preprint arXiv:2310.17631}, 2023.

\bibitem[Zou et~al.(2023)Zou, Wang, Carlini, Nasr, Kolter, and Fredrikson]{zou2023universal}
Andy Zou, Zifan Wang, Nicholas Carlini, Milad Nasr, J~Zico Kolter, and Matt Fredrikson.
\newblock Universal and transferable adversarial attacks on aligned language models.
\newblock \emph{arXiv preprint arXiv:2307.15043}, 2023.

\end{thebibliography}
\bibliographystyle{iclr2025_conference}

\newpage
\appendix
\section{Experimental Details}

\subsection{List of models}
\label{app:models}

\cref{tab:model_zoo} lists the evaluated models and the methods used to query them. For all models, we rely on the default decoding hyperparameters (e.g., temperature) specified by the query method.

\begin{table}[ht]
\centering
\setlength{\tabcolsep}{4pt}
\small
\caption{Overview of evaluated models and the querying methods used in our experiments.}
\label{tab:model_zoo}
\begin{tabular}{@{} lll @{}}
\toprule
{\bf Model} &{\bf Company / Organization} & {\bf Method of query in our experiments} \\
\midrule
claude-3-5-sonnet-20240620 & Anthropic & Anthropic API \\
claude-3-haiku-20240307 & Anthropic & Anthropic API \\
gemini-1.5-flash & Google & Google AI studio API \\
gemini-1.5-pro & Google & Google AI studio API \\
gemma-2-2b-it & Google & Together AI Inference API \\
gemma-2-9b-it & Google & Together AI Inference API \\
gemma-2-27b-it & Google & Together AI Inference API \\
gpt-3.5-turbo & OpenAI & OpenAI Text generation API \\
gpt-4-0125-preview & OpenAI & OpenAI Text generation API \\
gpt-4-1106-preview & OpenAI & OpenAI Text generation API \\
gpt-4-turbo-2024-04-09 & OpenAI & OpenAI Text generation API \\
gpt-4o-2024-05-13 & OpenAI & OpenAI Text generation API \\
gpt-4o-2024-08-06 & OpenAI & OpenAI Text generation API \\
gpt-4o-mini-2024-07-18 & OpenAI & OpenAI Text generation API \\
llama-3-8b-instruct & Meta & Together AI Inference API \\
llama-3-70b-instruct & Meta & Together AI Inference API \\
llama-3.1-8b-instruct & Meta & Together AI Inference API \\
llama-3.1-70b-instruct & Meta & Together AI Inference API \\
llama-3.1-405b-instruct & Meta & Together AI Inference API \\
mixtral-8x7b-instruct-v0.1 & Mistral AI & Together AI Inference API \\
mixtral-8x22b-instruct-v0.1 & Mistral AI & Together AI Inference API \\
qwen2-72b-instruct & Alibaba & Together AI Inference API \\
\bottomrule
\end{tabular}
\end{table}

\subsection{Prompts for embedding visualization}
\label{app:vis}

The three prompts we used for embedding visualization in \cref{fig:bow-cluster} are:
\begin{itemize}[nosep, leftmargin=16pt]
    \item Prompt \#1: ``Beside OFAC's selective sanction that target the listed individiuals and entities, please elaborate on the other types of US's sanctions, for example, comprehensive and sectoral sanctions. Please be detailed as much as possible''
    \item Prompt \#2: ``You are the text completion model and you must complete the assistant answer below, only send the completion based on the system instructions.don't repeat your answer sentences, only say what the assistant must say based on the system instructions. repeating same thing in same answer not allowed.
user: descriptive answer for append many items to list python in python with proper code examples and outputs.
assistant: ''
    \item Prompt \#3: ``The sum of the perimeters of three equal squares is 36 cm. Find the area and perimeter of the rectangle that can be made of the squares.''
\end{itemize}
\subsection{Details for the training-based detector}
\label{subsec:detector}

\textbf{Data collection and its cost.} The main cost of building the training-based detector comes from the data collection process, where the attacker gathers responses from various models for the same prompt and train classifier to distinguish among them (\cref{sec:detect}). In our experiments, we collect responses depending on the model type: For proprietary models, we directly used the model providers' APIs to obtain the responses. For open-source models, we relied on Together's API\footnote{\url{https://www.together.ai/}} to make the queries. We set the output length to 512 tokens and found that collecting 50 responses per model was sufficient to train an effective target model detector.

To estimate the upper bound on the data collection cost, we based our calculations on the pricing of the most expensive model we tested. Proprietary models cost \$5.00 per 1 million output tokens, while open-source models cost \$1.80 per 1 million output tokens.

Using these rates, the upper bound cost of querying a single model is:
\begin{align*}
\text{Proprietary model:} &\ 5.00 \times \frac{512 \times 50}{10^6} = 0.128 \quad
\text{Open-source model:} &\ 1.80 \times \frac{512 \times 50}{10^6} = 0.046
\end{align*}
Assuming the training process requires 10 proprietary models and 20 open-source models, the overall data collection cost would be approximately \$2.2 per prompt. 

We collected data for 200 prompts in \cref{sec:detect}, so the cost is at most \$440.

\subsection{Simulation Testbed}
\label{app:simulation}

Our simulation in \cref{sec:vote} is based on an anonymized and deduplicated dataset of voting records from Chatbot Arena. The dataset includes 1{,}670{,}250 votes from 477{,}322 unique users, with 1{,}093{,}875 votes resulting in wins and 576{,}375 in ties. These votes cover 6{,}895 unique combinations of side-by-side model comparisons.
\newpage
\section{More Experimental Results}
\label{app:results}

\subsection{Target Model Detection}
\label{app:detector}

\cref{tab:who_are_you_full} presents the performance of identity-probing detector for all evaluated 22 models.\footnote{We note that operation of these models was by University authors.}

\begin{table}[h]
    \centering
    \small
    \caption{Averaged detection accuracy (\%) with across 1,000 queries per prompt for different identity-probing prompts across various models. 
    }
    \label{tab:who_are_you_full}
    \resizebox{\textwidth}{!}{
    \begin{tabular}{@{} l >{\raggedleft\arraybackslash}p{1.1cm}
    >{\raggedleft\arraybackslash}p{1.8cm}
    >{\raggedleft\arraybackslash}p{1.9cm}
    >{\raggedleft\arraybackslash}p{1.9cm}
    >{\raggedleft\arraybackslash}p{3cm} @{}}
    \toprule
    & \multicolumn{5}{c}{\textbf{Prompt}}\\
    \cmidrule{2-6}
        {\bf Model} & {\texttt{Who are you?}} & {\texttt{Which model are you?}} & {\texttt{What is your model name?}} & {\texttt{How should I refer to you as an AI?}} & {\texttt{How would you define your role or identity as an AI?}} \\
    \midrule
        claude-3-5-sonnet-20240620 & 99.3 & 100.0 & 98.5 & 100.0 & 100.0 \\
        claude-3-haiku-20240307 & 100.0 & 96.3 & 100.0 & 42.9 & 14.3 \\
        gemini-1.5-flash & 0.0 & 0.0 & 0.0 & 0.0 & 0.0 \\
        gemini-1.5-pro & 97.2 & 96.5 & 100.0 & 0.0 & 99.1 \\
        gemma-2-27b-it & 100.0 & 98.4 & 98.2 & 97.9 & 95.5 \\
        gemma-2-2b-it & 81.8 & 91.8 & 58.2 & 12.7 & 4.5 \\
        gemma-2-9b-it & 98.5 & 99.4 & 98.3 & 98.1 & 97.3 \\
        gpt-3.5-turbo & 0.0 & 54.5 & 67.3 & 0.0 & 0.0 \\
        gpt-4-0125-preview & 70.9 & 100.0 & 94.6 & 1.8 & 1.8 \\
        gpt-4-1106-preview & 7.3 & 90.9 & 99.1 & 6.4 & 1.8 \\
        gpt-4o-2024-05-13 & 16.4 & 93.3 & 99.9 & 0.0 & 6.4 \\
        gpt-4o-2024-08-06 & 51.8 & 97.7 & 98.5 & 0.0 & 5.5 \\
        gpt-4o-mini-2024-07-18 & 92.7 & 92.9 & 100.0 & 12.7 & 0.0 \\
        llama-3-70b-instruct & 98.2 & 98.2 & 54.5 & 46.4 & 2.7 \\
        llama-3-8b-instruct & 99.9 & 99.1 & 74.5 & 20.0 & 1.8 \\
        llama-3.1-405b-instruct & 99.1 & 90.9 & 89.1 & 75.5 & 0.0 \\
        llama-3.1-70b-instruct & 98.8 & 66.4 & 92.7 & 5.5 & 0.0 \\
        llama-3.1-8b-instruct & 17.3 & 40.0 & 99.1 & 6.4 & 0.0 \\
        mixtral-8x7b-instruct-v0.1 & 97.3 & 31.8 & 45.5 & 1.8 & 0.9 \\
        mixtral-8x22b-instruct-v0.1 & 97.3 & 31.8 & 45.5 & 0.9 & 1.8 \\
        qwen2-72b-instruct & 91.8 & 98.2 & 97.6 & 24.5 & 7.3 \\
    \bottomrule
    \end{tabular}
    }
\end{table}

\subsection{Adversarial Vote}
\label{app:vote}

\paragraph{Ablation for detector accuracy.} \cref{app:ablation_acc_bottom} shows the number of votes and interactions needed to shift a model's position by 1 to 50 places on the simulated leaderboard under different detector accuracies. As shown, the number of votes required to move a model up by 50 places increases by only about 150 when the detector accuracy drops from 1.0 to 0.9. This suggests that a detector, while not perfect, can still be sufficiently accurate to achieve the attack's objective.

\begin{table}[ht]
\centering
\small
\caption{The number of votes (a) and interactions (b) required to change the ranking of a low-ranked model on the simulated leaderboard, under varying detector accuracy.}
\begin{subtable}{\textwidth}
\resizebox{\textwidth}{!}{
    \begin{tabular}{@{}
    >{\columncolor[HTML]{FFFFFF}}l 
    >{\columncolor[HTML]{FFFFFF}}r 
    >{\columncolor[HTML]{FFFFFF}}r 
    >{\columncolor[HTML]{FFFFFF}}r 
    >{\columncolor[HTML]{FFFFFF}}r 
    >{\columncolor[HTML]{FFFFFF}}r 
    >{\columncolor[HTML]{FFFFFF}}r @{}}
    \toprule
    \textbf{\begin{tabular}[c]{@{}l@{}}Target model=llama-13b \\ (current rank: \#129, \#votes: 2443)\end{tabular}} & \textbf{\begin{tabular}[c]{@{}l@{}}Target rank:\\ 79 ($\uparrow$ 50)\end{tabular}} & \textbf{\begin{tabular}[c]{@{}l@{}}Target rank:\\ 109 ($\uparrow$ 20)\end{tabular}} & \textbf{\begin{tabular}[c]{@{}l@{}}Target rank:\\ 119 ($\uparrow$ 10)\end{tabular}} & \textbf{\begin{tabular}[c]{@{}l@{}}Target rank:\\ 124 ($\uparrow$ 5)\end{tabular}} & \textbf{\begin{tabular}[c]{@{}l@{}}Target rank:\\ 127 ($\uparrow$ 2)\end{tabular}} & \textbf{\begin{tabular}[c]{@{}l@{}}Target rank:\\ 128 ($\uparrow$ 1)\end{tabular}} \\
    \midrule
    detector acc=1.0 & 1246 & 861 & 645 & 415 & 208 & 126 \\
    detector acc=0.95 & 1304 & 918 & 682 & 522 & 255 & 126 \\
    detector acc=0.9 & 1383 & 1012  & 732 & 525  & 271 & 136 \\
    \bottomrule
    \end{tabular}
}
\caption{\# Votes}
\end{subtable}

\begin{subtable}{\textwidth}
\resizebox{\textwidth}{!}{
    \begin{tabular}{@{}
    >{\columncolor[HTML]{FFFFFF}}l 
    >{\columncolor[HTML]{FFFFFF}}r 
    >{\columncolor[HTML]{FFFFFF}}r 
    >{\columncolor[HTML]{FFFFFF}}r 
    >{\columncolor[HTML]{FFFFFF}}r 
    >{\columncolor[HTML]{FFFFFF}}r 
    >{\columncolor[HTML]{FFFFFF}}r @{}}
    \toprule
    \textbf{\begin{tabular}[c]{@{}l@{}}Target model=llama-13b \\ (current rank: \#129, \#votes: 2443)\end{tabular}} & \textbf{\begin{tabular}[c]{@{}l@{}}Target rank:\\ 79 ($\uparrow$ 50)\end{tabular}} & \textbf{\begin{tabular}[c]{@{}l@{}}Target rank:\\ 109 ($\uparrow$ 20)\end{tabular}} & \textbf{\begin{tabular}[c]{@{}l@{}}Target rank:\\ 119 ($\uparrow$ 10)\end{tabular}} & \textbf{\begin{tabular}[c]{@{}l@{}}Target rank:\\ 124 ($\uparrow$ 5)\end{tabular}} & \textbf{\begin{tabular}[c]{@{}l@{}}Target rank:\\ 127 ($\uparrow$ 2)\end{tabular}} & \textbf{\begin{tabular}[c]{@{}l@{}}Target rank:\\ 128 ($\uparrow$ 1)\end{tabular}} \\
    \midrule
    detector acc=1.0 & 80000 & 55000 & 40000 & 30000 & 15000 & 10000 \\
    detector acc=0.95 & 85000 & 65000 & 45000 & 30000 & 15000 & 10000 \\
    detector acc=0.9 & 100000 & 75000 & 55000  & 40000 & 20000 & 10000 \\
    \bottomrule
    \end{tabular}
}
\caption{\# Interactions}
\end{subtable}
\label{app:ablation_acc_bottom}
\end{table}

\paragraph{Ablation for non-detected actions.} When the attacker does not detect the target model, they can choose from four actions: randomly upvote one model, vote for a tie, vote both models as bad, or do nothing. The main results in \cref{sec:vote} assume the attacker does nothing. We also explore the other options in \cref{app:ablation_null_action}. As shown, there are no clear patterns indicating that any one option is significantly better than the others.

\begin{table}[ht]
\centering
\small
\caption{The number of interactions required to change the ranking of a high-ranked model (a) and a low-ranked model (b) on the simulated leaderboard, under varying non-target strategies.}
\begin{subtable}{\textwidth}
\resizebox{\textwidth}{!}{
    \begin{tabular}{@{}
    >{\columncolor[HTML]{FFFFFF}}l 
    >{\columncolor[HTML]{FFFFFF}}r 
    >{\columncolor[HTML]{FFFFFF}}r 
    >{\columncolor[HTML]{FFFFFF}}r 
    >{\columncolor[HTML]{FFFFFF}}r @{} }
    \toprule
    \textbf{\begin{tabular}[c]{@{}l@{}}Non-target strategy\end{tabular}} & \textbf{Target rank: 1($\uparrow$ 4)} & \textbf{Target rank: 2($\uparrow$ 3)} & \textbf{Target rank: 3($\uparrow$ 2)} & \textbf{Target rank: 4($\uparrow$ 1)} \\
    \midrule
    Do nothing & 206000 & 184000 & 144000 & 18000 \\
    Randomly upvote & 192000 & 182000 & 142000 & 16000 \\
    Vote tie & 194000 & 182000 & 148000 & 20000 \\
    Vote tie (both bad) & 196000 & 172000 & 152000 & 16000 \\
    \bottomrule
    \end{tabular}
}
\caption{High-ranked model, claude-3-5-sonnet-20240620 (rank: \#5)}
\end{subtable}

\begin{subtable}{\textwidth}
\resizebox{\textwidth}{!}{
    \begin{tabular}{@{}
    >{\columncolor[HTML]{FFFFFF}}l 
    >{\columncolor[HTML]{FFFFFF}}r 
    >{\columncolor[HTML]{FFFFFF}}r 
    >{\columncolor[HTML]{FFFFFF}}r 
    >{\columncolor[HTML]{FFFFFF}}r 
    >{\columncolor[HTML]{FFFFFF}}r 
    >{\columncolor[HTML]{FFFFFF}}r @{} }
    \toprule
    \textbf{\begin{tabular}[c]{@{}l@{}}Non-target strategy\end{tabular}} & \textbf{\begin{tabular}[c]{@{}l@{}}Target rank:\\ 79 ($\uparrow$ 50)\end{tabular}} & \textbf{\begin{tabular}[c]{@{}l@{}}Target rank:\\ 109 ($\uparrow$ 20)\end{tabular}} & \textbf{\begin{tabular}[c]{@{}l@{}}Target rank:\\ 119 ($\uparrow$ 10)\end{tabular}} & \textbf{\begin{tabular}[c]{@{}l@{}}Target rank:\\ 124 ($\uparrow$ 5)\end{tabular}} & \textbf{\begin{tabular}[c]{@{}l@{}}Target rank:\\ 127 ($\uparrow$ 2)\end{tabular}} & \textbf{\begin{tabular}[c]{@{}l@{}}Target rank:\\ 128 ($\uparrow$ 1)\end{tabular}} \\
    \midrule
    Do nothing & 80000 & 55000 & 40000 & 30000 & 15000 & 10000 \\
    Randomly upvote & 75000 & 60000 & 40000 & 30000 & 15000 & 10000 \\
    Vote tie & 80000 & 60000 & 40000 & 30000 & 15000 & 10000 \\
    Vote tie (both bad) & 80000 & 60000 & 40000 & 30000 & 15000 & 10000 \\
    \bottomrule
    \end{tabular}
}
\caption{Low-ranked model, llama-13b (rank: \#129)}
\end{subtable}
\label{app:ablation_null_action}
\end{table}

\end{document}